\documentclass[11pt]{article}

\usepackage[final]{acl}
\usepackage{times}
\usepackage{latexsym}
\usepackage{amsmath}
\usepackage{amssymb}
\usepackage[T1]{fontenc}

\usepackage[utf8]{inputenc}

\usepackage{microtype}

\usepackage{inconsolata}

\usepackage{graphicx}

\usepackage{graphicx}

\usepackage{booktabs}
\usepackage{multirow}

\usepackage{xcolor}
\usepackage{caption}
\usepackage{amssymb}
\usepackage{amsmath}
\usepackage{subfigure}
\usepackage{subcaption}
\usepackage{color}
\usepackage[table]{xcolor}

\usepackage{hyperref}
\usepackage{url}
\usepackage{arydshln}
\usepackage{placeins}

\def\method{SEEK}

\title{\method{}: Steering LLM Reasoning for RAG via Internal Reasoning Sketches}

\author{%
  \textbf{Xinze Li}$^{1}$,
   \textbf{Yuqing Lan}$^{1}$,
  \textbf{Zhenghao Liu}$^{1}$\thanks{~~indicates corresponding author.},
  \textbf{Haidong Xin}$^{1}$,
  \textbf{Yukun Yan}$^{2}$\footnotemark[1],\\
  \textbf{Shuo Wang}$^{2}$,
  \textbf{Zheni Zeng}$^{3}$,
  \textbf{Sen Mei}$^{2}$,
  \textbf{Ge Yu}$^{1}$,
  \textbf{Maosong Sun}$^{2}$\\
  $^{1}$School of Computer Science and Engineering, Northeastern University, China\\
  $^{2}$Department of Computer Science and Technology, Institute for AI, Tsinghua University, China\\
  $^{3}$School of Intelligent Science and Technology, Nanjing University, China\\
}

\begin{document}
\maketitle
\begin{abstract}
Retrieval-Augmented Generation (RAG) enhances Large Language Models (LLMs) by incorporating external knowledge into the generation process. Benefiting from the reasoning capabilities of LLMs, existing methods have leveraged such capabilities to enable iterative knowledge acquisition and accumulation, thereby better supporting answer generation. However, as the reasoning trajectory grows, the accumulated knowledge and previously generated queries may interfere with subsequent retrieval decisions, resulting in sub-queries with repetitive intents and redundant knowledge acquisition. To address this issue, we propose \method{}, a sketch-guided knowledge acquisition framework for RAG. \method{} first prompts the LLM to construct a structured steering sketch for the given question. It consists of multiple groups of steering gists, with each gist followed by a slot for knowledge filling. Guided by these steering gists, \method{} iteratively retrieves and refines knowledge, and fills the corresponding slots to complete the sketch. The completed sketch is then used as contextual input for final answer generation. Experimental results show that \method{} achieves better performance than baseline models across multiple tasks. Further analyses demonstrate that \method{} can generate more diverse sub-queries, reduce redundant retrieval, and achieve a better balance between external knowledge utilization and internal knowledge conflict mitigation. All codes are available at \url{https://github.com/OpenBMB/PAGER}.
\end{abstract}


\section{Introduction}
Retrieval-Augmented Generation (RAG) improves the performance of Large Language Models (LLMs) by retrieving external documents to serve as supporting evidence~\citep{guu2020retrieval,LewisPPPKGKLYR020}. Early work has primarily focused on leveraging retrieval tools~\citep{bge_embedding} to perform a single-pass retrieval over an external corpus for a given query, using the retrieved knowledge as contextual input to guide LLMs in answer generation~\citep{izacard2023atlas,shi2024replug,xu2023lmgqs,liu2025autoencoding}. However, single-pass retrieval often provides insufficient information to fully answer a question, particularly in multi-hop reasoning scenarios~\citep{lin2025fishing,tangmultihop}, thereby limiting the effectiveness of RAG models in complex and multi-step question answering tasks.

\begin{figure}[!t] 
\centering
\includegraphics[width=\linewidth]{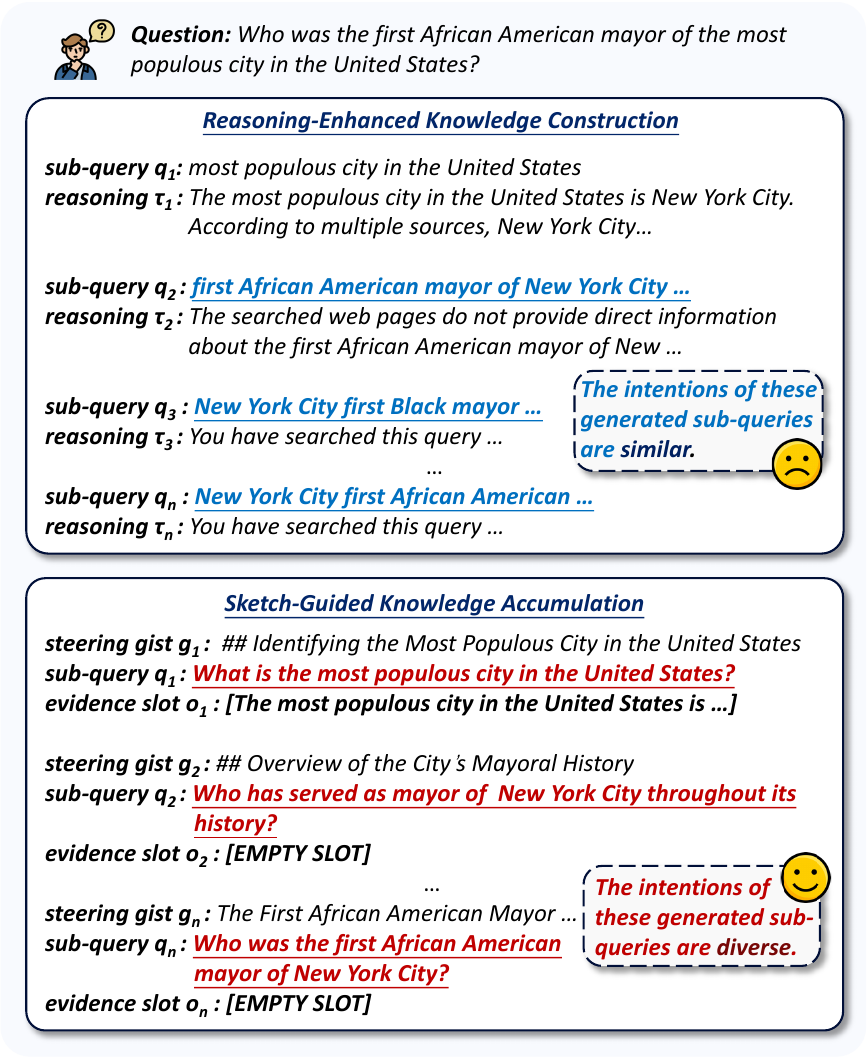}
\caption{Information Requirements of Different Reasoning-Guided RAG Models.} 
\label{fig:intro}
\end{figure}

Benefiting from the strong reasoning capabilities of LLMs~\cite{guo2025deepseek}, existing approaches seek to acquire sufficient external knowledge for RAG systems by tightly coupling retrieval with the reasoning process~\cite{li2025search}. Specifically, these methods prompt LLMs to reason progressively over a given question, identify missing knowledge from the accumulated context, and iteratively generate sub-queries to retrieve additional relevant information, thereby improving answer accuracy~\citep{wang2025deepnotenotecentricdeepretrievalaugmented,li2025search}.
However, as illustrated in Figure~\ref{fig:intro}, the reasoning process can be easily distracted by both previously retrieved knowledge and earlier generated queries, often leading to semantically similar sub-queries~\citep{rezaei2025vendi}. Consequently, LLMs may gradually lose track of the original knowledge requirements needed to answer the question, resulting in redundant retrieval and ineffective exploration of query-relevant knowledge as the reasoning trajectory grows longer.

To address this problem, we propose \method{}, a \textbf{S}k\textbf{E}tch-guid\textbf{E}d \textbf{K}nowledge acquisition framework that leverages self-generated sketches to steer LLM reasoning during iterative retrieval. Unlike prior approaches that directly accumulate retrieved knowledge, \method{} first prompts the LLM to construct a structured steering sketch for a given question based on its own parametric knowledge. The sketch is composed of multiple groups of steering gists, each corresponding to a distinct knowledge dimension potentially required for answering the question, while reserving dedicated slots for subsequent knowledge accumulation. Guided by these steering gists, the LLM iteratively retrieves and fills the corresponding slots with external knowledge, progressively completing the sketch. Finally, \method{} uses the completed sketch as contextual evidence to support the LLM in generating the final answer.


Experimental results demonstrate that \method{} consistently outperforms all baseline methods on knowledge-intensive tasks across diverse scenarios and backbone models, validating both its effectiveness and robustness. Further analysis reveals that the structured steering sketch constructed by \method{} effectively guides LLMs in acquiring, organizing, and integrating knowledge, thereby supporting more accurate question answering. Under the guidance of steering gists, \method{} enables LLMs to generate diverse sub-queries and retrieve more comprehensive, structured, and accurate knowledge, leading to higher information gains. Moreover, by leveraging the reasoning capability of LLMs to incorporate external knowledge into the steering sketch, \method{} effectively alleviates the knowledge conflict phenomenon~\cite{xie2024adaptive} in LLMs and facilitates more effective consolidation of external knowledge for answer generation.

\section{Related Work}





Large Language Models (LLMs)~\citep{yang2025qwen3,dubey2024llama} have demonstrated strong capabilities across a wide range of natural language processing tasks~\citep{he2021efficient}. However, LLMs typically suffer from hallucination, which can lead to incorrect responses~\citep{jiang2023active,xu2023recomp}. To mitigate this issue, existing studies have employed RAG models, which perform retrieval for the question to obtain relevant documents and incorporate them as input context, enabling LLMs to generate more accurate answers~\citep{LewisPPPKGKLYR020,guu2020retrieval}. However, performing a single retrieval step only using the original query makes it difficult to provide LLMs with comprehensive knowledge~\citep{lin2025fishing,tangmultihop}, especially in complex task scenarios~\citep{tangmultihop}.

To address these challenges, recent studies have explored query decomposition-based RAG frameworks to retrieve more comprehensive knowledge. Some approaches directly prompt LLMs to decompose the original query into multiple sub-queries, which are then used to retrieve broader and more diverse evidence~\citep{DBLP:journals/corr/abs-2305-18323, DBLP:conf/acl/AmmannGA25}. However, these methods lack mechanisms for dynamically refining sub-queries based on intermediate evidence, limiting their ability to iteratively acquire missing knowledge. Other studies investigate iterative retrieval methods that alternate between retrieval and reasoning, leveraging previously retrieved evidence to guide subsequent sub-query generation~\citep{trivedi2023interleaving,shao2023enhancing}. Nevertheless, these approaches often perform insufficient reasoning and holistic analysis over the accumulated evidence, making it difficult to effectively identify knowledge gaps and retrieve more accurate supporting information.

Recently, some studies have shifted toward adopting reasoning-guided knowledge acquisition methods for RAG models~\citep {DBLP:conf/acl/FangMM25, wang2025deepnotenotecentricdeepretrievalaugmented}. These methods leverage the reasoning capabilities of LLMs to iteratively analyze the accumulated knowledge to generate sub-queries for retrieving documents, and progressively integrate them into a reasoning trajectory~\citep{wang2024rat, li2025search}. However, the reasoning process of these methods can be disturbed by the progressively accumulated external knowledge and previously generated queries, leading to semantically similar sub-queries~\citep{rezaei2025vendi}. 

\begin{figure*}[t]
\centering
\includegraphics[width=\linewidth]{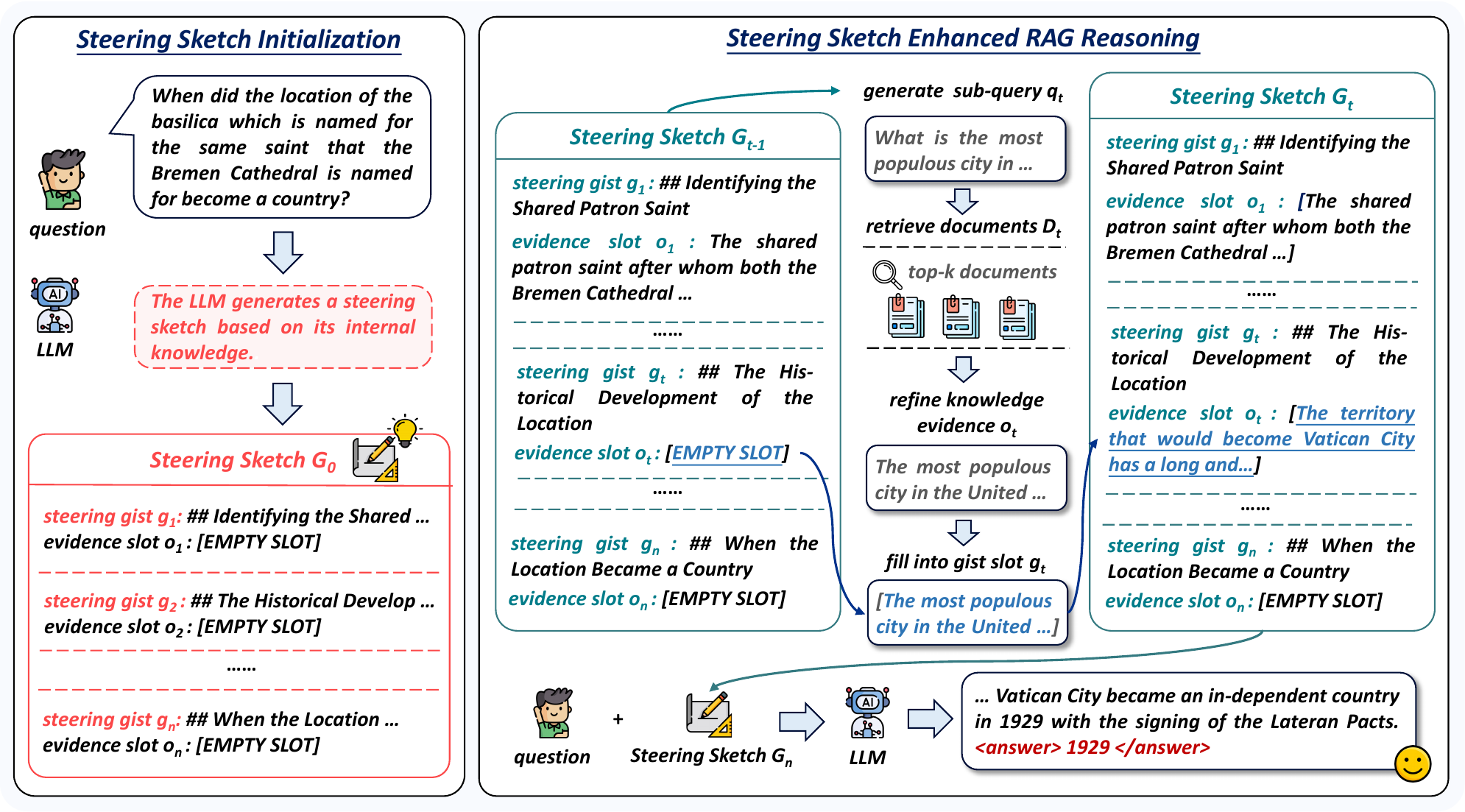}
\caption{The Illustration of Our Proposed \method{} Model.}
\label{fig:pipeline}
\end{figure*}
\section{Methodology}
\label{sec:method}
In this section, we present \method{}, a sketch-guided reasoning framework for effectively collecting knowledge to support the LLM in answering the given question. We first introduce the overall framework of reasoning-enhanced knowledge construction (Sec.~\ref{sec:3.1}). We then describe \method{} in detail, which leverages steering sketches to construct knowledge and thereby improve the effectiveness of RAG models (Sec.~\ref{sec:3.2}).

\subsection{RAG Modeling via Reasoning-Enhanced Knowledge Construction}
\label{sec:3.1}
To enable LLMs to solve complex questions, existing methods typically leverage the reasoning capabilities of LLMs to acquire sufficient external knowledge~\cite{li2025search}. Specifically, these methods iteratively generate sub-queries based on the current reasoning trajectory, retrieve relevant external knowledge, and progressively integrate the retrieved knowledge into it.

Given a question $q$, the model maintains a reasoning trajectory $R_{t-1}$ at the $t$-th step:
\begin{equation}
\small
R_{t-1} = (\tau_1, q_1, o_1, \ldots, \tau_{t-1}, q_{t-1}, o_{t-1}),
\end{equation}
where $\tau_i$ denotes an intermediate reasoning step that analyzes the knowledge accumulated in the current reasoning trajectory, $q_i$ denotes the sub-query generated for retrieval, and $o_i$ denotes the knowledge evidence derived by the model through reasoning over the retrieved knowledge.

At the $t$-th iteration, LLM $\mathcal{M}$ first reads the current reasoning trajectory $R_{t-1}$ and generates a new intermediate reasoning step $\tau_t$ with a sub-query $q_{t}$. The sub-query $q_t$ is used to retrieve the top-$k$ documents $D_{t}$ from the external corpus:
\begin{equation}\label{eq:retriever}
\small
D_{t} = \text{Retriever}(q_{t}, k),
\end{equation}
where $\text{Retriever}$ denotes the retrieval tool. The LLM further reasons over the retrieved knowledge $D_t$ and derives knowledge evidence $o_t$:
\begin{equation}
\small
o_t = \mathcal{M}(R_{t-1}, \tau_t, q_t, D_t).
\end{equation}
By appending $(\tau_t, q_t, o_t)$ to $R_{t-1}$, the reasoning trajectory $R_{t-1}$ is updated to $R_{t}$.
This process is iteratively repeated until the accumulated knowledge is sufficient to answer the question or the maximum number of iterations is reached, yielding the final reasoning trajectory $R_T$. Finally, the $R_T$ is regarded as the knowledge context to answer the question $q$:
\begin{equation}\label{eq:qa}
\small
y = \mathcal{M}(q, R_T),
\end{equation}
where $y$ is the final answer generated by the LLM $\mathcal{M}$. Despite effective, as the reasoning trajectory grows longer, the accumulated information and previously generated queries also gradually accumulate, which interferes with the model’s reasoning process. The generated sub-queries $q_t$ tend to exhibit repetitive retrieval intents, resulting in redundant retrieval.

\subsection{Sketch-Guided Knowledge Acquisition During LLM Reasoning}
\label{sec:3.2}
To further enable the RAG model to generate more accurate and diverse sub-queries for iterative retrieval, \method{} leverages a structured steering sketch to guide knowledge acquisition. Unlike existing reasoning-enhanced RAG models, \method{} initializes a steering sketch derived from the LLM's own reasoning process, which contains multiple steering gists and fillable slots. The model iteratively acquires knowledge under the guidance of these steering gists and fills the knowledge evidence into the slots, thereby producing a completed sketch. This completed sketch is used as contextual input to guide the model in generating the final answer.


\paragraph{Steering Sketch Enhanced RAG Reasoning.} Given a question $q$, \method{} first prompts the LLM $\mathcal{M}$ to analyze the question, identify the key knowledge aspects required for answering it, and initialize a steering sketch $G_0$ without relying on external documents:
\begin{equation}
\small
G_0 =\mathcal{M}(q),
\end{equation}
where $G_0$ consists of a set of $n$ logically progressive steering gists $g_i$ to guide the knowledge acquisition, and each steering gist is paired with an empty slot $[\ ]_i$ for filling in the acquired knowledge evidence:
\begin{equation}
\small
G_0 = (g_1, [\ ]_1, \ldots, g_n, [\ ]_n).
\end{equation}

\method{} starts from the steering sketch state $G_0$ and iteratively acquires knowledge under the guidance of each steering gist $g_i$ to sequentially fill the blank slots $[\ ]_{1:n}$. At the $t$-th iteration step, the $(t-1)$-th sketch state is represented as follows:
\begin{equation}
\small
G_{t-1} = (g_1, [o]_1, \ldots, g_{t-1}, [o]_{t-1}, g_t, [\ ]_t, \ldots, g_n, [\ ]_n),
\end{equation}
where $o_i$ denotes the refined knowledge evidence already filled into the $i$-th slot. Then, under the guidance of steering gists, the model retrieves and refines the knowledge into the knowledge evidence $o_t$, which is used to fill the $t$-th slot to update $G_{t-1}$ to $G_t$:
\begin{equation}
\small
G_{t} = (g_1, [o]_1, \ldots, g_t, [o]_t, g_{t+1}, [\ ]_{t+1},\ldots, g_n, [\ ]_n),
\end{equation}
This iterative process continues until, after $n$ iterations, all $n$ blank slots in the sketch are filled with knowledge evidence, resulting in the completed sketch $G_n$. Finally, different from Eq~\ref{eq:qa}, \method{} uses the completed sketch $G_n$ as knowledge context to generate the final answer $y$:
\begin{equation}
\small
y = \mathcal{M}(q, G_n).
\end{equation}

\paragraph{Steering Sketch Updating.} To update the $(t-1)$-th steering sketch state $G_{t-1}$ to $G_t$ at the $t$-th iteration, \method{} first prompts the LLM to generate a sub-query $q_t$ with a reasoning step $\tau_t$, under the guidance of the $t$-th steering gist in $G_{t-1}$:
\begin{equation}
\small
\tau_t, q_t = \mathcal{M}(q, G_{t-1}).
\end{equation}
Unlike the reasoning trajectories introduced in Sec.~\ref{sec:3.1}, $\tau_t$ and $q_t$ are used exclusively during the sketch updating stage and are not persistently retained in the context for subsequent knowledge acquisition. This design ensures that the sketch remains focused on essential knowledge and steering signals, without being cluttered by distracting intermediate information. \method{} then leverages the generated sub-query $q_t$ to retrieve the top-$k$ documents $D_t$ (Eq.~\ref{eq:retriever}), which are subsequently refined into the knowledge evidence $o_t$:
\begin{equation}
\small
o_t = \mathcal{M}(q, q_t, D_t, G_{t-1}).
\end{equation}
Finally, \method{} obtains the $t$-th sketch state $G_{t}$ by filling the knowledge evidence $o_t$ into the $t$-th slot $[\ ]_t$ of $G_{t-1}$:
\begin{equation}\small
o_t \xrightarrow{\mathrm{fill}} [\ ]_t.
\end{equation}

\section{Experimental Methodology}
In this section, we describe the datasets, evaluation metrics, and baselines.

\paragraph{Dataset.} 
Following previous work~\citep{li2025search,song2025r1}, we evaluate our method on both multi-hop and single-hop QA benchmarks. Specifically, we select HotpotQA~\citep{yang2018hotpotqa}, 2WikiMultiHopQA~\citep{ho2020constructing}, MuSiQue~\citep{trivedi2022musique}, and Bamboogle~\cite{press2023measuring} for multi-hop tasks, while using NQ~\citep{kwiatkowski2019natural} and AmbigQA~\citep{min2020ambigqa} for single-hop tasks. For evaluation, we randomly sample 2,000 instances from the dev set of each dataset, except for Bamboogle~\cite{press2023measuring}, where we utilize the test set (125 instances) due to its limited size.

\begin{table*}[t]
\centering
\small
\resizebox{\linewidth}{!}{
\begin{tabular}{lccccccc}
\toprule
\textbf{Methods} & \textbf{HotpotQA} & \textbf{2WikiMQA} & \textbf{MuSiQue}& \textbf{Bamboogle} & \textbf{NQ} & \textbf{AmbigQA} & \textbf{Avg.} \\
\midrule
\rowcolor{gray!8}\multicolumn{8}{l}{\textbf{\textit{Qwen3-32B}}} \\
\midrule
Vanilla LLM& 28.6 &31.8 &8.2 & 44.0 &36.2 & 32.9 &30.3\\
Vanilla RAG & 42.9 &36.5 &13.5& 52.0 & 54.6 &54.8 &42.4\\
StructRAG~\citeyearpar{DBLP:conf/iclr/LiC0L0T0H0L25} &42.5 &31.5&12.8&44.8&55.1&54.5&40.2\\
Iter-RetGen~\citeyearpar{shao2023enhancing} &43.2 &40.2 &14.5  &37.6 & \textbf{56.5} &55.4&41.2\\
RAT~\citeyearpar{wang2024rat} &43.7 &45.0&16.1&\underline{55.2} &53.2&55.5&\underline{44.8} \\
ReWOO~\citeyearpar{DBLP:journals/corr/abs-2305-18323} &43.6 &50.7 &14.1 &51.2 &54.4 &52.1 &44.4 \\
Search-o1~\citeyearpar{li2025search} &47.6  &47.0 &\underline{22.9}  &33.6 & 49.2 &51.5 &42.0\\
DeepNote~\citeyearpar{wang2025deepnotenotecentricdeepretrievalaugmented}   & \underline{48.4} & \underline{47.2} & 17.2 &39.2 & \underline{55.6} & \underline{55.9}  &43.9 \\
\method{} & \textbf{50.6} &\textbf{57.4} &\textbf{23.0}&\textbf{62.4} & \textbf{56.5}& \textbf{56.4}&\textbf{51.1} \\
\midrule
\rowcolor{gray!8}\multicolumn{8}{l}{\textbf{\textit{Llama3.1-70B-Instruct}}} \\
\midrule
Vanilla LLM& 37.7 &41.5 &14.1 & 57.6&50.3 &48.8 & 41.7 \\
Vanilla RAG & 48.2 &41.9 &19.1 &56.8 &55.7 &57.0 & 46.5 \\
StructRAG~\citeyearpar{DBLP:conf/iclr/LiC0L0T0H0L25} &49.6&44.1&19.4&57.6&\textbf{57.4}&58.2&47.7\\
Iter-RetGen~\citeyearpar{shao2023enhancing} &41.1 &34.5 & 11.9&32.8 &53.0 &52.3 & 37.6\\
RAT~\citeyearpar{wang2024rat} & 48.9 &40.2&20.8&\underline{58.4}&53.6&56.4&46.2\\
ReWOO~\citeyearpar{DBLP:journals/corr/abs-2305-18323} &45.2&44.7&17.0&50.4&54.3&54.0&44.3\\
Search-o1~\citeyearpar{li2025search} &50.2 &\textbf{55.4} &\underline{22.9}  &\underline{58.4} &51.6  &53.8&48.3\\
DeepNote~\citeyearpar{wang2025deepnotenotecentricdeepretrievalaugmented} &\underline{51.7} &48.9 &20.8 &\underline{58.4} &\underline{56.6} & \underline{58.4}&\underline{49.1} \\
\method{} &\textbf{52.4} &\underline{54.9} &\textbf{24.3} & \textbf{62.4}&56.4 & \textbf{60.0}&\textbf{51.7} \\
\bottomrule
\end{tabular}
}
\caption{Overall Performance of Different RAG Models. The \textbf{best} and \underline{second best} results are highlighted.}
\label{tab:overall}
\end{table*}

\paragraph{Baselines.}
In our experiments, we compare \method{} with multiple baseline models, including a Vanilla LLM, one-pass retrieval RAG models, and iterative retrieval RAG models. For the Vanilla LLM, we directly feed the query to the LLM and ask it to generate the answer. For one-pass retrieval RAG models, we adopt Vanilla RAG, StructRAG~\citep{DBLP:conf/iclr/LiC0L0T0H0L25}, and ReWOO~\citep{DBLP:journals/corr/abs-2305-18323}. Vanilla RAG utilizes retrieved documents as contextual input to assist the LLM in answering the question. StructRAG designs a router to refine documents into structured knowledge representations as input context. ReWOO generates all sub-queries for the given question in a single pass and executes them in parallel to retrieve external knowledge. For iterative retrieval RAG models, we employ Iter-RetGen~\citep{shao2023enhancing}, RAT~\citep{wang2024rat}, Search-o1~\citep{li2025search}, and DeepNote~\citep{wang2025deepnotenotecentricdeepretrievalaugmented} to accumulate knowledge. Specifically, Iter-RetGen interleaves retrieval with the generation process, utilizing intermediate generated content to guide subsequent retrieval. RAT directly prompts the LLM to generate a CoT and iteratively retrieves external knowledge to refine each reasoning step in the COT. DeepNote compresses retrieved knowledge into a note and iteratively acquires additional information to update it. Search-o1 leverages the reasoning capabilities of LLMs to guide iterative knowledge acquisition and applies the Reason-in-Document mechanism to refine the knowledge.

\paragraph{Implementation Details.} In our experiments, we employ Qwen3-32B~\citep{yang2025qwen3} and Llama-3.1-70B-Instruct~\citep{dubey2024llama} as backbone models. We follow existing work~\citep{sun2025rearter,song2025r1} to utilize Cover Exact Match as the evaluation metric. We follow FlashRAG~\citep{jin2025flashrag} and UltraRAG~\citep{chen2025ultrarag} to use Wikipedia as the retrieval corpus, and adopt Qwen3-Embedding-0.6B~\citep{zhang2025qwen3} as the embedding model for retrieval. For each retrieval step, we retain the top-5 retrieved documents for all RAG models.

\section{Experimental Results}
In this section, we first evaluate the performance of \method{} across different models and datasets. Subsequently, we conduct ablation studies to analyze the effectiveness of different functional components in \method{}. Then, we investigate the effectiveness of steering gists for knowledge construction. Finally, we introduce the quality and utilization of accumulated knowledge.

\begin{table*}[t]
\centering
\small
\resizebox{\linewidth}{!}{
\begin{tabular}{lccccccc}
\toprule
\textbf{Methods} & \textbf{HotpotQA} & \textbf{2WikiMQA} & \textbf{MuSiQue}& \textbf{Bamboogle} & \textbf{NQ} & \textbf{AmbigQA} & \textbf{Avg.} \\
\midrule
\rowcolor{gray!8}\multicolumn{8}{l}{\textbf{\textit{Qwen3-32B}}} \\
\midrule

\method{} (Parallel Filling) &45.9 &43.8 &18.8 &59.2 &\underline{56.4} &\textbf{57.0} &46.9\\
\method{} (Plan) & 47.0  &50.7&21.2&59.2 &54.6&54.5&47.9\\

\method{} (Update) &\underline{49.9} &\underline{56.6} &\textbf{23.3} &60.0 &55.2 &56.0 &\underline{50.2} \\
\hdashline
\method{} & \textbf{50.6} &\textbf{57.4} &\underline{23.0} &\textbf{62.4} & \textbf{56.5}& \underline{56.4}& \textbf{51.1}\\
\quad w/o IterRetrieval &44.8 &42.2 &16.9 &52.0 &54.6 &55.1 &44.3  \\
\quad w/o Initialization &46.2&45.2&15.9&48.0&55.2&55.9&44.4 \\
\midrule
\rowcolor{gray!8}\multicolumn{8}{l}{\textbf{\textit{Llama3.1-70B-Instruct}}} \\
\midrule

\method{} (Parallel Filling) 
&47.4&43.4&18.6&58.4&\textbf{57.2}&\underline{59.5}&47.4\\
\method{} (Plan) &  52.3 &54.3 &23.9 &56.0 &57.0 &59.5& 50.5 \\

\method{} (Update) & \textbf{52.9} &\textbf{57.6} &\textbf{25.8} &59.4 &\underline{57.0} &59.2 &\textbf{52.0} \\
\hdashline
\method{} &\underline{52.4} &\underline{54.9} &\underline{24.3} &\textbf{62.4} &56.4 &\textbf{60.0} & \underline{51.7}\\
\quad w/o IterRetrieval &47.6 &41.6 &18.1 &56.0 &56.3 &58.1 & 46.3 \\
\quad w/o Initialization &51.1 & 50.5 &21.4 &\underline{60.0} &55.9 &57.9 &49.5 \\
\bottomrule
\end{tabular}
}
\caption{Ablation Study. The \textbf{best} and \underline{second best} results are highlighted.}
\label{tab:ablation}
\end{table*}

\subsection{Overall Performance}
As shown in Table~\ref{tab:overall}, we compare the overall performance of \method{} with various baseline methods across a range of tasks.

Overall, \method{} demonstrates its effectiveness by outperforming all baseline models, achieving improvements exceeding 2\%. Notably, \method{} consistently shows improvements across various tasks and backbone LLMs, underscoring its robust generalization ability. When compared with ReWOO, \method{} achieves an average performance improvement of over 5\%, indicating that, instead of generating all sub-queries at once for parallel knowledge acquisition, \method{} enables the model to continuously interact with external knowledge, dynamically adjust the retrieval direction, and acquire more comprehensive knowledge. Furthermore, compared with iterative retrieval-based methods such as Iter-RetGen, \method{} achieves an improvement of over 9\%, highlighting its role in more effectively acquiring and organizing retrieved knowledge. 
In addition, \method{} outperforms DeepNote and Search-o1, indicating that, under the guidance of the steering gists, the model can better reason about and analyze the knowledge gaps to acquire more comprehensive knowledge, thereby providing stronger support for LLMs in answering questions.


\subsection{Ablation Study}\label{ablation}
In this subsection, we conduct ablation studies to evaluate the effectiveness of different components in \method{}.

In the experiments, we compare \method{} with five ablated models. \method{} (Parallel Filling) simultaneously generates sub-queries for all missing slots in the initial sketch, performs parallel retrieval, and then refines the retrieved documents to fill the corresponding slots. \method{} (Plan) uses the sketch as a preceding plan to guide the iterative retrieval and knowledge refinement process, where the refined knowledge obtained at each step is no longer filled into slots but sequentially concatenated until the accumulated knowledge is sufficient to answer the question. \method{} (Update) dynamically updates the sketch structure during the iterative sketch completion process, including deleting, adding, and updating the steering gists in the sketch. \method{} (w/o IterRetrieval) performs a single-pass retrieval based on the given query and fills the initial sketch using the retrieved documents. \method{} (w/o Initialization) iteratively refines the retrieved documents into concise summaries and continuously concatenates them until the aggregated summaries are sufficient to answer the question. This variant removes the stage of generating the sketch with the LLM. Additional implementation details are provided in Appendix~\ref{app:ablation}.

As shown in Table~\ref{tab:ablation}, \method{} consistently outperforms \method{} (Parallel Filling), indicating that \method{} can iteratively generate more targeted sub-queries based on the previous sketch state to retrieve more comprehensive information. Moreover, \method{} outperforms \method{} (Plan), indicating that the sketch is not merely a simple plan but serves as a structured steering signal to guide LLM reasoning, thereby enabling more effective knowledge acquisition and organization. Furthermore, \method{} achieves comparable or even better performance than \method{} (Update), while requiring lower inference time. This suggests that the sketch constructed by \method{} provides soft guidance for the knowledge acquisition process and does not strictly constrain the direction of query generation. In contrast, modifying the sketch structure may disrupt the logical coherence of such guidance. In addition, compared with \method{} (w/o IterRetrieval), \method{} exhibits consistent performance gains across different datasets and backbone models, indicating that iterative retrieval can incorporate more essential knowledge to better answer the given query. Finally, compared with \method{} (w/o Initialization), \method{} achieves further improvements, highlighting the critical role of the sketch generated by leveraging the reasoning capability of LLMs.


\begin{figure}[t]
  \centering
  \subfigure[Sub-query Similarity.]{
    \includegraphics[width=0.47\linewidth]{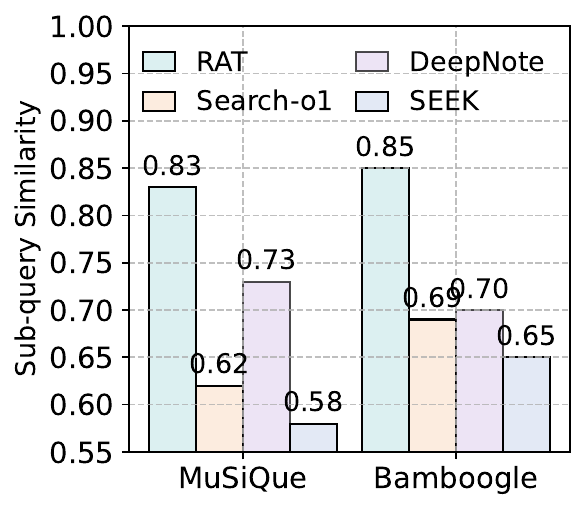}\label{fig:subquery_similarity}
    
  }
    \subfigure[Document Overlap.]{
    \includegraphics[width=0.46\linewidth]{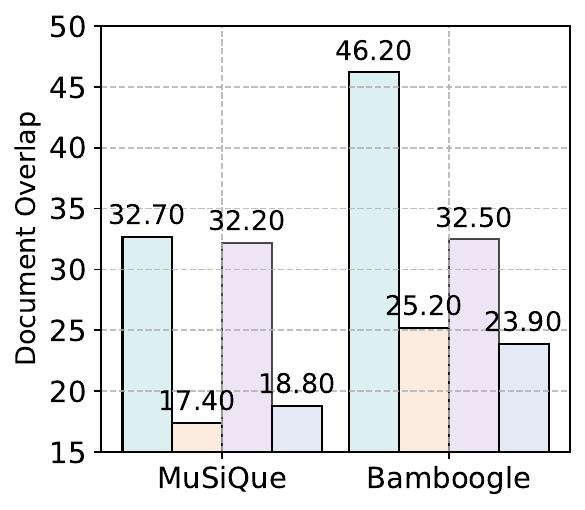}\label{fig:overlap}
    
  }
  \caption{Diversity of Sub-queries Generated by Different Methods.}
  \label{fig:query_quality}
\end{figure}

\begin{figure}[t]
  \centering
  \subfigure[MuSiQue.]{
    \includegraphics[width=0.46\linewidth]{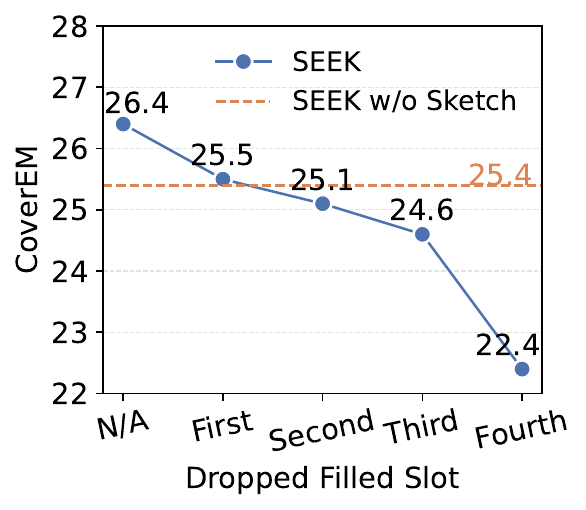}
    
    \label{fig:drop_musique}
  }
  \hfill
  \subfigure[Bamboogle.]{
    \includegraphics[width=0.46\linewidth]{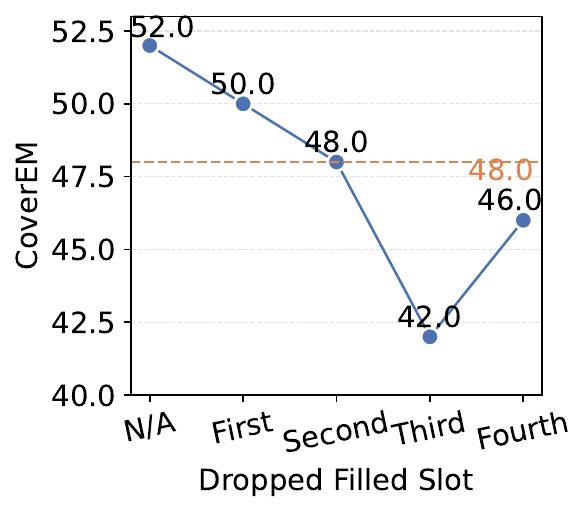}
    \label{fig:bamboogle}
  }
  \caption{Slot Ablation Studies on the Completed Sketch. ``N/A'' denotes the completed sketch with no filled slots removed. ``First'', ``Second'', ``Third'', and ``Fourth'' denote the variants in which the First, Second, Third, and Fourth filled slots are removed, respectively. SEEK (w/o Sketch) denotes a variant where all steering gists are removed from the completed sketch.
}
  \label{fig:section_drop}
\end{figure}
\subsection{Effectiveness of the Steering Sketch for Knowledge Construction}
In this section, we investigate the effectiveness of steering gists in the knowledge construction process of \method{}. In this experiment, we use Qwen3-32B as the backbone model and evaluate on the MuSiQue and Bamboogle datasets.


\paragraph{Diversity of the Generated Sub-queries.}
Figure~\ref{fig:query_quality} evaluates the diversity of generated sub-queries from two perspectives, query-intent diversity and retrieval redundancy. Specifically, we first use Qwen3-Embedding-0.6B to compute the semantic similarity among sub-queries generated at different iterations, aiming to evaluate whether different methods tend to generate queries with similar intents. We then compute the Jaccard similarity between the documents retrieved at different retrieval steps to measure the overlap among documents retrieved by different sub-queries.

As shown in Figure~\ref{fig:subquery_similarity}, the results show that \method{} achieves the lowest sub-query similarity on both datasets, indicating that steering gists can guide the LLM to generate more diverse sub-queries toward different knowledge aspects. In addition, as shown in Figure~\ref{fig:overlap}, \method{} produces substantially lower document overlap than RAT and DeepNote, suggesting that the steering sketch can effectively reduce redundant retrieval. These findings demonstrate that \method{} better alleviates sub-query intention convergence and facilitates the acquisition of more comprehensive knowledge.

\paragraph{Effectiveness of Completed Knowledge in Different Slots of the Sketch.}
As shown in Figure~\ref{fig:section_drop}, we further analyze the effectiveness of the knowledge filled in different slots of the sketch. We collect completed sketches containing four slots as seed sketches. As shown in Appendix~\ref{Statistics_of_page_slots}, the sketches constructed by Qwen3-32B predominantly contain four slots. Then, we remove the first, second, third, and fourth filled slots, respectively, to construct different incomplete sketches. Finally, we feed these four types of incomplete sketches into the model to evaluate their performance.

Overall, compared with the completed sketch \method{} (N/A), removing any filled slot from the sketch leads to performance degradation, indicating that all filled knowledge is necessary to support LLMs in answering the query. As the removed filled slot shifts from the first to the fourth slot, the model performance drops more substantially, suggesting that the knowledge filled in later slots is more critical for question answering. One possible reason is that our iterative completion method tends to acquire increasingly necessary knowledge in later slots, which aligns with the reasoning process of LLMs, where later reasoning steps often involve deeper inference to answer the query. Furthermore, compared with \method{} (N/A), \method{} (w/o Sketch) also exhibits a performance decline. This observation further indicates that the structured steering sketch plays a critical role in guiding the knowledge construction process, thereby validating the effectiveness of the sketch initialization module.







\begin{figure}[t]
  \centering
    \subfigure[Information Gain.]{
    \includegraphics[width=0.48\linewidth]{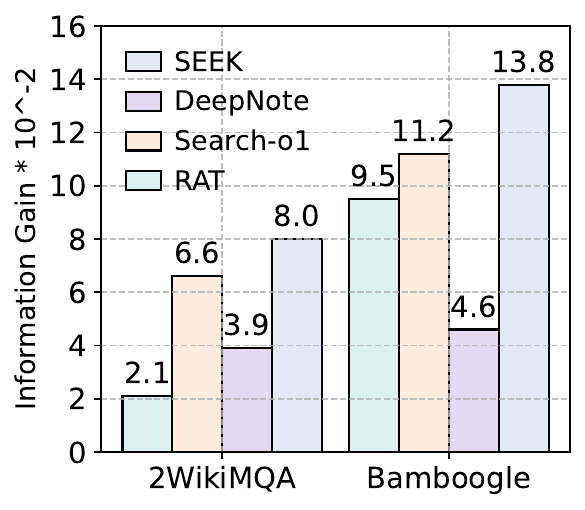}
    \label{fig:infor}
  }
  \subfigure[GPT-5.1 Evaluation Score.]{
    \includegraphics[width=0.42\linewidth]{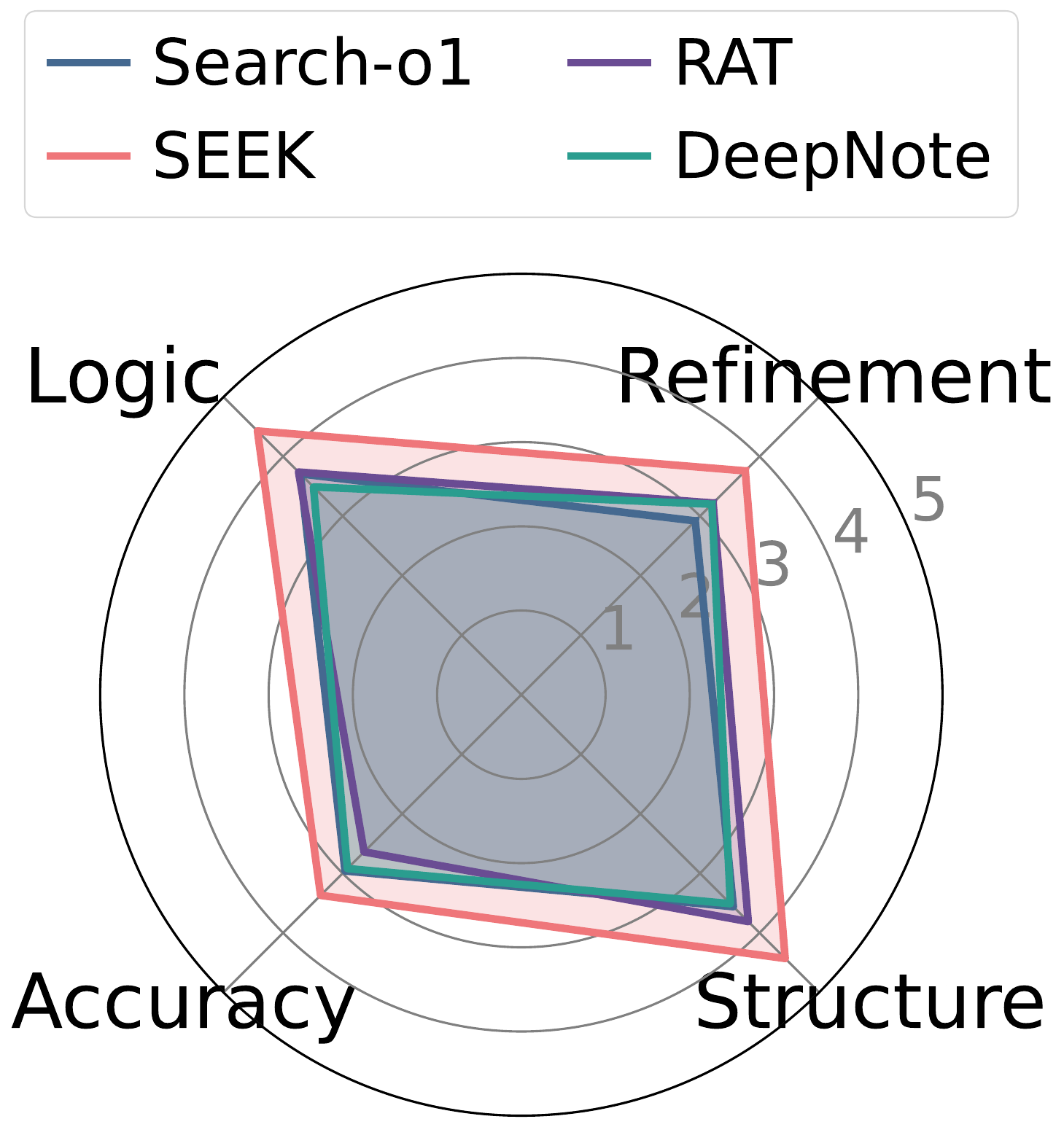}
    \label{fig:gpt}
  }
  \caption{The Quality of Knowledge Constructed by Different Methods.}
  \label{fig:page_quality}
\end{figure}

\subsection{Quality and Utilization of Accumulated Knowledge}
In this section, we investigate the quality and utilization of accumulated knowledge in different methods. We adopt Qwen3-32B as the backbone model for all experiments.

As shown in Figure~\ref{fig:page_quality}, we evaluate the effectiveness of knowledge construction across different RAG models. As shown in Figure~\ref{fig:infor}, we first compute the information gain of the accumulated knowledge of different RAG models~\citep{DBLP:journals/corr/abs-2509-12765}, where information gain quantifies the contribution of the knowledge context fed into the model to correct answer generation. Details are provided in Appendix~\ref{app:infogain}. The results show that \method{} achieves higher information gain than other methods, which indicates that the structured sketches incorporate more relevant knowledge to guide LLMs toward generating accurate answers. We further evaluate the quality of the knowledge context generated by different models. As shown in Figure~\ref{fig:gpt}, we employ a strong closed-source LLM, GPT-5.1, as the evaluator. Using the prompt templates provided in Appendix~\ref{app:prompt}, we assess the knowledge context along four dimensions: accuracy, logicality, structure, and degree of knowledge refinement, with each dimension rated on a scale from 0 to 5. The evaluation results demonstrate that \method{} consistently outperforms other models across all dimensions, with particularly notable advantages in structure and logical consistency. These findings further validate the effectiveness of the steering sketches constructed by \method{} in knowledge acquisition and accumulation, enabling the LLM to acquire the necessary knowledge under the guidance of steering gists.

\begin{table}[t]
\centering
\small
\begin{tabular}{lccc}
\toprule
\textbf{Methods}  &\textbf{2WikiMQA}  &\textbf{Bamboogle} &\textbf{HotpotQA}\\
\midrule
\rowcolor{gray!8}\multicolumn{4}{l}{\textbf{\textit{Knowledge Conflict}}} \\
\midrule
Vanilla LLM& 100.0 & 100.0  & 100.0  \\
Vanilla RAG & 61.1 & 90.9 & 82.2\\
DeepNote &64.5  &58.2  &82.2\\ 
RAT &72.8 &\textbf{92.7} & 84.9\\
Search-o1 &56.3 &52.7 &82.6 \\
\method{} &\textbf{73.4}  &83.6  & \textbf{87.6} \\ 
\midrule
\rowcolor{gray!8}\multicolumn{4}{l}{\textbf{\textit{Knowledge Utilization}}} \\
\midrule
Vanilla LLM& 0.0 & 0.0  & 0.0  \\
Vanilla RAG &25.1 &21.4 &27.2\\
DeepNote &39.2 &24.3 &34.9\\ 
RAT &32.0 &25.7 &27.2\\
Search-o1 &42.7 &18.6 & 33.5\\ 
\method{} &\textbf{49.9}  &\textbf{45.7} & \textbf{35.8} \\
\bottomrule
\end{tabular}
\caption{Performance of Different RAG Models under Different Testing Scenarios.}
\label{tab:conflict}
\end{table}
Next, we evaluate the impact of different ways of acquiring and accumulating knowledge on knowledge conflicts and knowledge utilization in LLMs. As shown in Table~\ref{tab:conflict}, we design two evaluation scenarios: knowledge conflict and knowledge utilization. 
For the knowledge conflict scenario, we select samples from the evaluation dataset where the LLM can generate correct answers solely based on its parametric knowledge, aiming to assess the denoising capability of RAG systems. For the knowledge utilization scenario, we construct the evaluation set by selecting samples where the LLM must rely on external knowledge to generate the answer. 
The evaluation results show that \method{} and RAT perform better in the knowledge conflict scenario, indicating that using the reasoning process derived from the LLM's own parametric knowledge to guide knowledge acquisition can alleviate conflicts between internal and external knowledge. 
In contrast, DeepNote and Search-o1 outperform RAT in the knowledge utilization setting, suggesting that incorporating more retrieved knowledge provides greater potential for correcting factual errors in memorized knowledge. 
Notably, \method{} achieves the best performance in both the knowledge utilization and conflict scenarios. These results suggest that, by combining structured steering sketches with reasoning-guided knowledge acquisition, \method{} can achieve a more targeted balance between mitigating knowledge conflicts and enhancing knowledge utilization.

\section{Conclusion}
This paper presents \method{}, a sketch-guided knowledge acquisition framework for RAG. By constructing a structured steering sketch and iteratively filling its slots with retrieved evidence, \method{} organizes external knowledge into a completed sketch that better supports answer generation. Experimental results show that \method{} consistently outperforms baselines across multiple tasks. 
\section*{Limitations}

Although \method{} achieves superior performance across multiple datasets, the sketch filling process introduces additional latency. Specifically, to ensure the logical coherence and completeness of the constructed sketch, \method{} adopts an iterative slot-filling mechanism. This iterative process inevitably incurs extra computational overhead, thereby increasing inference latency. Moreover, we also explore a parallel filling variant of \method{} to mitigate this issue. However, experimental results show that the iterative design remains necessary for maintaining the effectiveness of \method{} when answering multi-hop QA queries. Therefore, the trade-off between effectiveness and efficiency remains a critical challenge for fully deploying \method{} in real-world QA scenarios.

\bibliography{custom_new}

@inproceedings{trivedi2023interleaving,
 address = {Toronto, Canada},
 author = {Trivedi, Harsh  and
Balasubramanian, Niranjan  and
Khot, Tushar  and
Sabharwal, Ashish},
 booktitle = {Proceedings of the 61st Annual Meeting of the Association for Computational Linguistics (Volume 1: Long Papers)},
 doi = {10.18653/v1/2023.acl-long.557},
 editor = {Rogers, Anna  and
Boyd-Graber, Jordan  and
Okazaki, Naoaki},
 pages = {10014--10037},
 publisher = {Association for Computational Linguistics},
 title = {Interleaving Retrieval with Chain-of-Thought Reasoning for Knowledge-Intensive Multi-Step Questions},
 url = {https://aclanthology.org/2023.acl-long.557},
 year = {2023}
}

@misc{wang2025deepnotenotecentricdeepretrievalaugmented,
 author = {Ruobing Wang and Qingfei Zhao and Yukun Yan and Daren Zha and Yuxuan Chen and Shi Yu and Zhenghao Liu and Yixuan Wang and Shuo Wang and Xu Han and Zhiyuan Liu and Maosong Sun},
 journal = {ArXiv preprint},
 title = {DeepNote: Note-Centric Deep Retrieval-Augmented Generation},
 url = {https://arxiv.org/abs/2410.08821},
 volume = {abs/2410.08821},
 year = {2024}
}

@inproceedings{he2021efficient,
 address = {Online and Punta Cana, Dominican Republic},
 author = {He, Junxian  and
Neubig, Graham  and
Berg-Kirkpatrick, Taylor},
 booktitle = {Proceedings of the 2021 Conference on Empirical Methods in Natural Language Processing},
 doi = {10.18653/v1/2021.emnlp-main.461},
 editor = {Moens, Marie-Francine  and
Huang, Xuanjing  and
Specia, Lucia  and
Yih, Scott Wen-tau},
 pages = {5703--5714},
 publisher = {Association for Computational Linguistics},
 title = {Efficient Nearest Neighbor Language Models},
 url = {https://aclanthology.org/2021.emnlp-main.461},
 year = {2021}
}

@article{li2025search,
 author = {Li, Xiaoxi and Dong, Guanting and Jin, Jiajie and Zhang, Yuyao and Zhou, Yujia and Zhu, Yutao and Zhang, Peitian and Dou, Zhicheng},
 journal = {ArXiv preprint},
 title = {Search-o1: Agentic search-enhanced large reasoning models},
 url = {https://arxiv.org/abs/2501.05366},
 volume = {abs/2501.05366},
 year = {2025}
}

@inproceedings{jiang2023active,
 address = {Singapore},
 author = {Jiang, Zhengbao  and
Xu, Frank  and
Gao, Luyu  and
Sun, Zhiqing  and
Liu, Qian  and
Dwivedi-Yu, Jane  and
Yang, Yiming  and
Callan, Jamie  and
Neubig, Graham},
 booktitle = {Proceedings of the 2023 Conference on Empirical Methods in Natural Language Processing},
 doi = {10.18653/v1/2023.emnlp-main.495},
 editor = {Bouamor, Houda  and
Pino, Juan  and
Bali, Kalika},
 pages = {7969--7992},
 publisher = {Association for Computational Linguistics},
 title = {Active Retrieval Augmented Generation},
 url = {https://aclanthology.org/2023.emnlp-main.495},
 year = {2023}
}

@article{xu2023recomp,
 author = {Xu, Fangyuan and Shi, Weijia and Choi, Eunsol},
 journal = {ArXiv preprint},
 title = {Recomp: Improving retrieval-augmented lms with compression and selective augmentation},
 url = {https://arxiv.org/abs/2310.04408},
 volume = {abs/2310.04408},
 year = {2023}
}

@inproceedings{yang2018hotpotqa,
 address = {Brussels, Belgium},
 author = {Yang, Zhilin  and
Qi, Peng  and
Zhang, Saizheng  and
Bengio, Yoshua  and
Cohen, William  and
Salakhutdinov, Ruslan  and
Manning, Christopher D.},
 booktitle = {Proceedings of the 2018 Conference on Empirical Methods in Natural Language Processing},
 doi = {10.18653/v1/D18-1259},
 editor = {Riloff, Ellen  and
Chiang, David  and
Hockenmaier, Julia  and
Tsujii, Jun{'}ichi},
 pages = {2369--2380},
 publisher = {Association for Computational Linguistics},
 title = {{H}otpot{QA}: A Dataset for Diverse, Explainable Multi-hop Question Answering},
 url = {https://aclanthology.org/D18-1259},
 year = {2018}
}

@inproceedings{shao2023enhancing,
 address = {Singapore},
 author = {Shao, Zhihong  and
Gong, Yeyun  and
Shen, Yelong  and
Huang, Minlie  and
Duan, Nan  and
Chen, Weizhu},
 booktitle = {Findings of the Association for Computational Linguistics: EMNLP 2023},
 doi = {10.18653/v1/2023.findings-emnlp.620},
 editor = {Bouamor, Houda  and
Pino, Juan  and
Bali, Kalika},
 pages = {9248--9274},
 publisher = {Association for Computational Linguistics},
 title = {Enhancing Retrieval-Augmented Large Language Models with Iterative Retrieval-Generation Synergy},
 url = {https://aclanthology.org/2023.findings-emnlp.620},
 year = {2023}
}

@article{song2025r1,
 author = {Song, Huatong and Jiang, Jinhao and Min, Yingqian and Chen, Jie and Chen, Zhipeng and Zhao, Wayne Xin and Fang, Lei and Wen, Ji-Rong},
 journal = {ArXiv preprint},
 title = {R1-searcher: Incentivizing the search capability in llms via reinforcement learning},
 url = {https://arxiv.org/abs/2503.05592},
 volume = {abs/2503.05592},
 year = {2025}
}

@inproceedings{ho2020constructing,
 address = {Barcelona, Spain (Online)},
 author = {Ho, Xanh  and
Duong Nguyen, Anh-Khoa  and
Sugawara, Saku  and
Aizawa, Akiko},
 booktitle = {Proceedings of the 28th International Conference on Computational Linguistics},
 doi = {10.18653/v1/2020.coling-main.580},
 editor = {Scott, Donia  and
Bel, Nuria  and
Zong, Chengqing},
 pages = {6609--6625},
 publisher = {International Committee on Computational Linguistics},
 title = {Constructing A Multi-hop {QA} Dataset for Comprehensive Evaluation of Reasoning Steps},
 url = {https://aclanthology.org/2020.coling-main.580},
 year = {2020}
}

@article{trivedi2022musique,
 address = {Cambridge, MA},
 author = {Trivedi, Harsh  and
Balasubramanian, Niranjan  and
Khot, Tushar  and
Sabharwal, Ashish},
 doi = {10.1162/tacl_a_00475},
 editor = {Roark, Brian  and
Nenkova, Ani},
 journal = {Transactions of the Association for Computational Linguistics},
 pages = {539--554},
 publisher = {MIT Press},
 title = {{M}u{S}i{Q}ue: Multihop Questions via Single-hop Question Composition},
 url = {https://aclanthology.org/2022.tacl-1.31},
 volume = {10},
 year = {2022}
}

@inproceedings{press2023measuring,
 address = {Singapore},
 author = {Press, Ofir  and
Zhang, Muru  and
Min, Sewon  and
Schmidt, Ludwig  and
Smith, Noah  and
Lewis, Mike},
 booktitle = {Findings of the Association for Computational Linguistics: EMNLP 2023},
 doi = {10.18653/v1/2023.findings-emnlp.378},
 editor = {Bouamor, Houda  and
Pino, Juan  and
Bali, Kalika},
 pages = {5687--5711},
 publisher = {Association for Computational Linguistics},
 title = {Measuring and Narrowing the Compositionality Gap in Language Models},
 url = {https://aclanthology.org/2023.findings-emnlp.378},
 year = {2023}
}

@article{sun2025rearter,
 author = {Sun, Zhongxiang and Wang, Qipeng and Yu, Weijie and Zang, Xiaoxue and Zheng, Kai and Xu, Jun and Zhang, Xiao and Yang, Song and Li, Han},
 journal = {ArXiv preprint},
 title = {Rearter: Retrieval-augmented reasoning with trustworthy process rewarding},
 url = {https://arxiv.org/abs/2501.07861},
 volume = {abs/2501.07861},
 year = {2025}
}

@misc{bge_embedding,
 author = {Shitao Xiao and Zheng Liu and Peitian Zhang and Niklas Muennighoff},
 journal = {ArXiv preprint},
 title = {C-Pack: Packaged Resources To Advance General Chinese Embedding},
 url = {https://arxiv.org/abs/2309.07597},
 volume = {abs/2309.07597},
 year = {2023}
}

@article{yang2025qwen3,
 author = {Yang, An and Li, Anfeng and Yang, Baosong and Zhang, Beichen and Hui, Binyuan and Zheng, Bo and Yu, Bowen and Gao, Chang and Huang, Chengen and Lv, Chenxu and others},
 journal = {ArXiv preprint},
 title = {Qwen3 technical report},
 url = {https://arxiv.org/abs/2505.09388},
 volume = {abs/2505.09388},
 year = {2025}
}

@inproceedings{jin2025flashrag,
 author = {Jin, Jiajie and Zhu, Yutao and Dou, Zhicheng and Dong, Guanting and Yang, Xinyu and Zhang, Chenghao and Zhao, Tong and Yang, Zhao and Wen, Ji-Rong},
 booktitle = {Companion Proceedings of the ACM on Web Conference 2025},
 pages = {737--740},
 title = {Flashrag: A modular toolkit for efficient retrieval-augmented generation research},
 year = {2025}
}

@inproceedings{LewisPPPKGKLYR020,
  author       = {Patrick Lewis and
                  Ethan Perez and
                  Aleksandra Piktus and
                  Fabio Petroni and
                  Vladimir Karpukhin and
                  Naman Goyal and
                  Heinrich K{\"{u}}ttler and
                  Mike Lewis and
                  Wen{-}tau Yih and
                  Tim Rockt{\"{a}}schel and
                  Sebastian Riedel and
                  Douwe Kiela},
  title        = {Retrieval-Augmented Generation for Knowledge-Intensive {NLP} Tasks},
  booktitle    = {Advances in Neural Information Processing Systems 33: Annual Conference
                  on Neural Information Processing Systems 2020, NeurIPS 2020, December
                  6-12, 2020, virtual},
  year         = {2020},
  url          = {https://proceedings.neurips.cc/paper/2020/hash/6b493230205f780e1bc26945df7481e5-Abstract.html},
}

@article{wang2024rat,
  title={Rat: Retrieval augmented thoughts elicit context-aware reasoning in long-horizon generation},
  author={Wang, Zihao and Liu, Anji and Lin, Haowei and Li, Jiaqi and Ma, Xiaojian and Liang, Yitao},
  journal={arXiv preprint arXiv:2403.05313},
  year={2024}
}

@article{dubey2024llama,
  title={The llama 3 herd of models},
  author={Dubey, Abhimanyu and Jauhri, Abhinav and Pandey, Abhinav and Kadian, Abhishek and Al-Dahle, Ahmad and Letman, Aiesha and Mathur, Akhil and Schelten, Alan and Yang, Amy and Fan, Angela and others},
  journal={arXiv e-prints},
  pages={arXiv--2407},
  year={2024}
}

@article{zhang2025qwen3,
  title={Qwen3 Embedding: Advancing Text Embedding and Reranking Through Foundation Models},
  author={Zhang, Yanzhao and Li, Mingxin and Long, Dingkun and Zhang, Xin and Lin, Huan and Yang, Baosong and Xie, Pengjun and Yang, An and Liu, Dayiheng and Lin, Junyang and others},
  journal={arXiv preprint arXiv:2506.05176},
  year={2025}
}

@article{kwiatkowski2019natural,
  title={Natural questions: a benchmark for question answering research},
  author={Kwiatkowski, Tom and Palomaki, Jennimaria and Redfield, Olivia and Collins, Michael and Parikh, Ankur and Alberti, Chris and Epstein, Danielle and Polosukhin, Illia and Devlin, Jacob and Lee, Kenton and others},
  journal={Transactions of the Association for Computational Linguistics},
  volume={7},
  pages={453--466},
  year={2019},
  publisher={MIT Press One Rogers Street, Cambridge, MA 02142-1209, USA journals-info~…}
}

@inproceedings{min2020ambigqa,
  title={AmbigQA: Answering Ambiguous Open-domain Questions},
  author={Min, Sewon and Michael, Julian and Hajishirzi, Hannaneh and Zettlemoyer, Luke},
  booktitle={Proceedings of the 2020 Conference on Empirical Methods in Natural Language Processing (EMNLP)},
  pages={5783--5797},
  year={2020}
}

@inproceedings{guu2020retrieval,
  title={Retrieval augmented language model pre-training},
  author={Guu, Kelvin and Lee, Kenton and Tung, Zora and Pasupat, Panupong and Chang, Mingwei},
  booktitle={International conference on machine learning},
  pages={3929--3938},
  year={2020},
  organization={PMLR}
}

@inproceedings{DBLP:conf/iclr/LiC0L0T0H0L25,
  author       = {Zhuoqun Li and
                  Xuanang Chen and
                  Haiyang Yu and
                  Hongyu Lin and
                  Yaojie Lu and
                  Qiaoyu Tang and
                  Fei Huang and
                  Xianpei Han and
                  Le Sun and
                  Yongbin Li},
  title        = {StructRAG: Boosting Knowledge Intensive Reasoning of LLMs via Inference-time
                  Hybrid Information Structurization},
  booktitle    = {The Thirteenth International Conference on Learning Representations},
  year         = {2025},
}

@inproceedings{xu2023lmgqs,
  title={LMGQS: A large-scale dataset for query-focused summarization},
  author={Xu, Ruochen and Wang, Song and Liu, Yang and Wang, Shuohang and Xu, Yichong and Iter, Dan and He, Pengcheng and Zhu, Chenguang and Zeng, Michael},
  booktitle={Findings of the Association for Computational Linguistics: EMNLP 2023},
  pages={14764--14776},
  year={2023}
}

@article{DBLP:journals/corr/abs-2509-12765,
  author       = {Zihan Wang and
                  Zihan Liang and
                  Zhou Shao and
                  Yufei Ma and
                  Huangyu Dai and
                  Ben Chen and
                  Lingtao Mao and
                  Chenyi Lei and
                  Yuqing Ding and
                  Han Li},
  title        = {InfoGain-RAG: Boosting Retrieval-Augmented Generation via Document
                  Information Gain-based Reranking and Filtering},
  journal      = {arXiv},
  year         = {2025},
}

@article{liu2025autoencoding,
    author = {Liu, Xin and Zhao, Runsong and Huang, Pengcheng and Liu, Xinyu and Xiao, Junyi and Xiao, Chunyang and Xiao, Tong and Gao, Shengxiang and Yu, Zhengtao and Zhu, Jingbo},
    journal = {ArXiv preprint},
    title = {Autoencoding-free context compression for llms via contextual semantic anchors},
    url = {https://arxiv.org/abs/2510.08907},
    volume = {abs/2510.08907},
    year = {2025}
}

@article{DBLP:journals/corr/abs-2305-18323,
    author = {Binfeng Xu and
Zhiyuan Peng and
Bowen Lei and
Subhabrata Mukherjee and
Yuchen Liu and
Dongkuan Xu},
    journal = {ArXiv preprint},
    title = {ReWOO: Decoupling Reasoning from Observations for Efficient Augmented
Language Models},
    url = {https://doi.org/10.48550/arXiv.2305.18323},
    year = {2023}
}

@article{rezaei2025vendi,
    author = {Rezaei, Mohammad Reza and Dieng, Adji Bousso},
    journal = {ArXiv preprint},
    title = {Vendi-rag: Adaptively trading-off diversity and quality significantly improves retrieval augmented generation with llms},
    url = {https://arxiv.org/abs/2502.11228},
    volume = {abs/2502.11228},
    year = {2025}
}

@article{izacard2023atlas,
    author = {Izacard, Gautier and Lewis, Patrick and Lomeli, Maria and Hosseini, Lucas and Petroni, Fabio and Schick, Timo and Dwivedi-Yu, Jane and Joulin, Armand and Riedel, Sebastian and Grave, Edouard},
    journal = {Journal of Machine Learning Research},
    number = {251},
    pages = {1--43},
    title = {Atlas: Few-shot learning with retrieval augmented language models},
    volume = {24},
    year = {2023}
}

@inproceedings{shi2024replug,
    author = {Shi, Weijia  and
Min, Sewon  and
Yasunaga, Michihiro  and
Seo, Minjoon  and
James, Richard  and
Lewis, Mike  and
Zettlemoyer, Luke  and
Yih, Wen-tau},
    booktitle = {Proceedings of NAACL},
    editor = {Duh, Kevin  and
Gomez, Helena  and
Bethard, Steven},
    pages = {8371--8384},
    title = {{REPLUG}: Retrieval-Augmented Black-Box Language Models},
    url = {https://aclanthology.org/2024.naacl-long.463},
    year = {2024}
}

@article{lin2025fishing,
    author = {Lin, Huifeng and Su, Gang and Liang, Jintao and Wu, You and Zhao, Rui and Li, Ziyue},
    journal = {ArXiv preprint},
    title = {Fishing for Answers: Exploring One-shot vs. Iterative Retrieval Strategies for Retrieval Augmented Generation},
    url = {https://arxiv.org/abs/2509.04820},
    volume = {abs/2509.04820},
    year = {2025}
}

@inproceedings{tangmultihop,
    author = {Tang, Yixuan and Yang, Yi},
    booktitle = {First Conference on Language Modeling},
    title = {MultiHop-RAG: Benchmarking Retrieval-Augmented Generation for Multi-Hop Queries},
    year = {2024}
}

@article{guo2025deepseek,
    author = {Guo, Daya and Yang, Dejian and Zhang, Haowei and Song, Junxiao and Wang, Peiyi and Zhu, Qihao and Xu, Runxin and Zhang, Ruoyu and Ma, Shirong and Bi, Xiao and others},
    journal = {ArXiv preprint},
    title = {Deepseek-r1: Incentivizing reasoning capability in llms via reinforcement learning},
    url = {https://arxiv.org/abs/2501.12948},
    volume = {abs/2501.12948},
    year = {2025}
}

@inproceedings{DBLP:conf/acl/AmmannGA25,
    author = {Paul J. L. Ammann and
Jonas Golde and
Alan Akbik},
    booktitle = {Proceedings of ACL},
    pages = {497--507},
    title = {Question Decomposition for Retrieval-Augmented Generation},
    url = {https://doi.org/10.18653/v1/2025.acl-srw.32},
    year = {2025}
}

@inproceedings{DBLP:conf/acl/FangMM25,
    author = {Jinyuan Fang and
Zaiqiao Meng and
Craig MacDonald},
    booktitle = {Proceedings of ACL},
    pages = {18969--18985},
    title = {KiRAG: Knowledge-Driven Iterative Retriever for Enhancing Retrieval-Augmented
Generation},
    url = {https://aclanthology.org/2025.acl-long.929/},
    year = {2025}
}

@article{chen2025ultrarag,
    author = {Chen, Yuxuan and Guo, Dewen and Mei, Sen and Li, Xinze and Chen, Hao and Li, Yishan and Wang, Yixuan and Tang, Chaoyue and Wang, Ruobing and Wu, Dingjun and others},
    journal = {ArXiv preprint},
    title = {UltraRAG: A Modular and Automated Toolkit for Adaptive Retrieval-Augmented Generation},
    url = {https://arxiv.org/abs/2504.08761},
    volume = {abs/2504.08761},
    year = {2025}
}

@inproceedings{xie2024adaptive,
  title={Adaptive chameleon or stubborn sloth: Revealing the behavior of large language models in knowledge conflicts},
  author={Xie, Jian and Zhang, Kai and Chen, Jiangjie and Lou, Renze and Su, Yu},
  booktitle={International Conference on Learning Representations},
  volume={2024},
  pages={35623--35646},
  year={2024}
}

\clearpage
\appendix

\section{Appendix}
\FloatBarrier

\subsection{License}
We show the licenses of the datasets that we use. All of these datasets are allowed for academic use under their respective licenses and agreements: MuSiQue and HotpotQA (CC-BY-4.0 License); 2WikiMQA (Apache 2.0 License); Bamboogle (MIT License); NQ and AmbigQA (CC BY-SA 3.0 License).

\begin{figure}[t] 
\centering
\includegraphics[width=\linewidth]{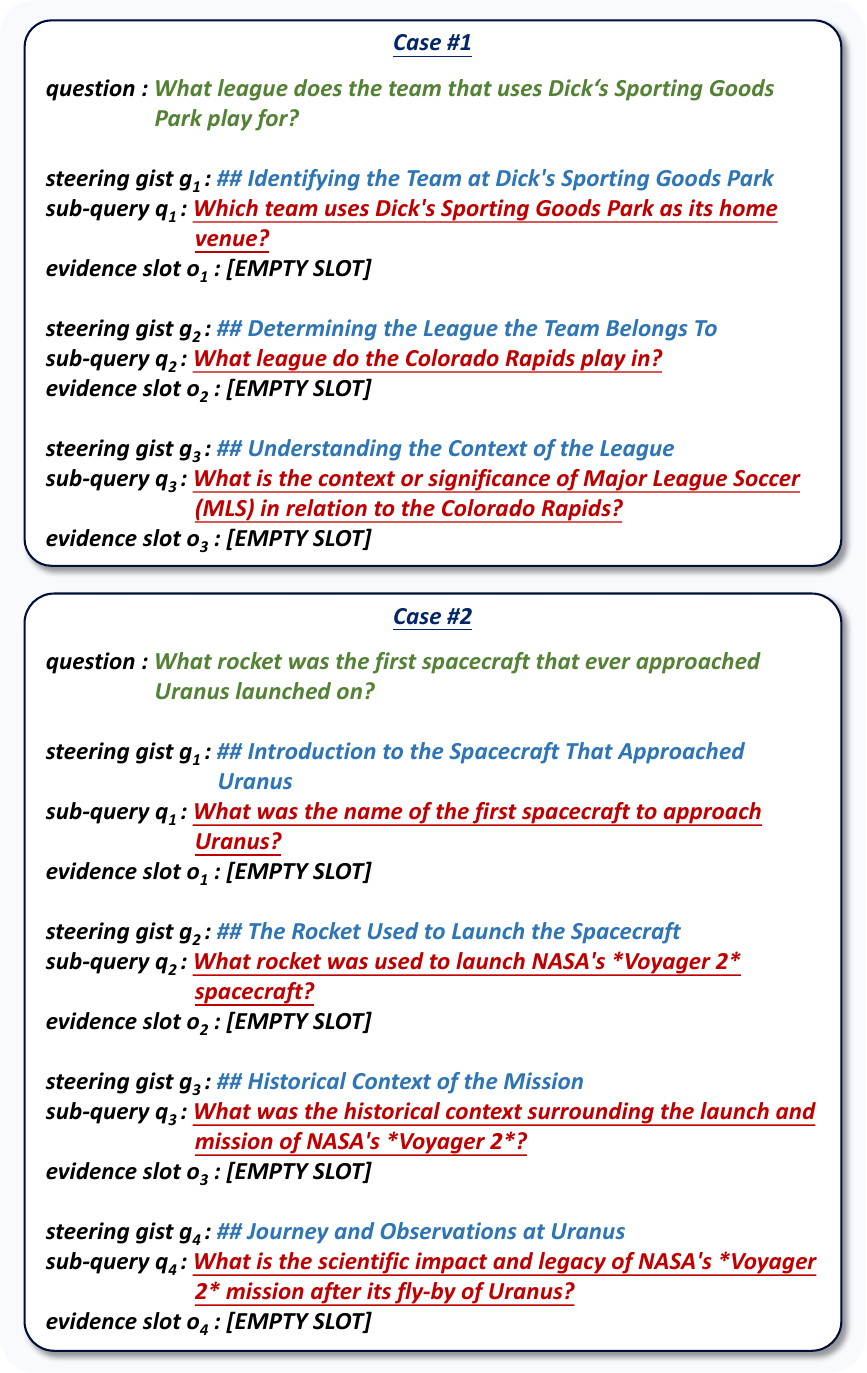}
\caption{Case Studies of the Difference Between Steering Gists and Generated Sub-queries. The steering gists provide structured guidance for knowledge acquisition, while the generated sub-queries are dynamically produced based on both the current gist and the evolving sketch state.} 
\label{fig:app:frame_query}
\end{figure}
\subsection{Differences Between Steering Gists and Generated Sub-queries} 
In this section, we present cases in Figure~\ref{fig:app:frame_query} to illustrate the distinction between the steering gists in the sketch and the generated sub-queries.

In the first case, each steering gist in the initialized sketch is concise and closely related to the question, without extending beyond the original entities and knowledge scope of the question. However, the generated sub-queries often include additional entity information that is not explicitly mentioned in the corresponding steering gists. For example, the sub-query generated under the guidance of the second steering gist includes the additional entity ``Colorado Rapids'', derived from the evidence provided for the first gist, although it does not appear in the second gist, ``Determining the League the Team Belongs To''. This indicates that \method{} generates sub-queries based on the current sketch state, rather than being strictly constrained by the textual topic of the current steering gist.

Furthermore, the new entity ``Major League Soccer'' in the third generated sub-query, together with the more detailed inquiry about the relationship between the league and the team, goes beyond the surface scope of the third steering gist and differs substantially from its original wording. This further shows that the model dynamically adjusts the intent and scope of each sub-query during sketch updating. In the second case, the iteratively generated sub-queries also introduce entities and relations that are not explicitly present in the steering gists, and such information becomes increasingly enriched as the sketch is progressively completed. These observations suggest that steering gists mainly provide a structured outline and soft guidance for knowledge acquisition, rather than rigidly determining the semantics of the generated sub-queries.

\subsection{Ablation Study Details} \label{app:ablation}
For the ablation variant \method{} (Update), this variant adaptively updates the steering sketch during the sketch completion process of \method{}. Specifically, \method{} (Update) introduces an additional step into the iterative knowledge completion process of \method{}, where the LLM is prompted to adjust the current sketch after the knowledge evidence has been filled into the corresponding slot. The adjustment process includes removing redundant or unnecessary slots, inserting new slots where appropriate, deleting irrelevant content from filled slots, and resolving conflicting information. This variant can be regarded as an exploration of the upper bound of the dynamic modeling capability of \method{}.

\begin{table}[t]
\centering
\small
\begin{tabular}{lccc}
\toprule
\textbf{Method} & \textbf{HotpotQA} & \textbf{MuSiQue} & \textbf{2WikiMQA} \\
\midrule
RAT       & 36.2 & 11.4 & 36.6 \\
DeepNote  & 46.8 & 15.9 & 41.5 \\
Search-o1 & 46.5 & 20.5 & 32.0 \\
\method{}     & 47.2 & 19.8 & 46.3 \\
\bottomrule
\end{tabular}
\caption{Experimental Results with BGE-Base-v1.5 as the Retrieval Model.}
\label{tab:bge_retriever}
\end{table}
\subsection{Additional Experiments with Different Retrievers} 
To further evaluate whether the effectiveness of \method{} depends on a specific retrieval model, we conduct additional experiments by replacing the original Qwen3-Embedding-0.6B retriever with BGE-Base-v1.5. The results are reported in Table~\ref{tab:bge_retriever}. We observe that \method{} continues to achieve stable improvements over all baseline methods under the new retriever. These results suggest that the performance improvement of \method{} is not tied to a particular embedding model, but exhibits robustness to different retrieval backbones.

\begin{table}[t]
\centering
\small
\begin{tabular}{lccc}
\toprule
\textbf{Method} & \textbf{HotpotQA} & \textbf{MuSiQue} & \textbf{2WikiMQA} \\
\midrule
RAT       & 57.60 & 29.00 & 45.00 \\
DeepNote  & 63.12 & 32.18 & 51.62 \\
Search-o1 & 64.55 & 37.20 & 58.70 \\
\method{} & 68.85 & 43.70 & 58.80 \\
\bottomrule
\end{tabular}
\caption{Maximum Recall@5 of Sub-queries Generated by Different Methods.}
\label{tab:subquery_recall}
\end{table}

\subsection{Sub-query Recall across Different Methods}
To directly evaluate whether the sub-queries generated by \method{} are more effective at the retrieval level, we further compute the maximum sub-query Recall@5 for different methods. Specifically, for each question, we first calculate Recall@5 for each generated sub-query, and then select the highest score among all sub-queries as the maximum Recall@5. This metric measures whether the model can generate at least one effective sub-query during the multi-step retrieval process to retrieve documents containing the ground-truth answer. We regard retrieved documents containing the ground-truth answer as an approximation of ground-truth evidence. This allows us to more directly examine whether the performance improvement of \method{} comes from more effective query planning, rather than only from subsequent knowledge integration. As shown in Table~\ref{tab:subquery_recall}, \method{} achieves higher maximum Recall@5 than the compared methods, indicating that the steering sketch can guide the model to generate more effective sub-queries for retrieving answer-relevant evidence.

\begin{table*}[t]
\centering
\small
\resizebox{\linewidth}{!}{
\begin{tabular}{lccccccc}
\toprule
\textbf{Methods} & \textbf{HotpotQA} & \textbf{2WikiMQA} & \textbf{MuSiQue}& \textbf{Bamboogle} & \textbf{NQ} & \textbf{AmbigQA} & \textbf{Avg.} \\
\midrule
\rowcolor{gray!8}\multicolumn{8}{l}{\textbf{\textit{GLM-4.5}}} \\
\midrule
Vanilla LLM&39.0 &57.0 &15.5 &62.4&58.0 &53.0 &47.5\\
Vanilla RAG &51.0 &63.0 &20.0 &62.4 &69.5 &62.5 & 54.7\\
DeepNote~\citeyearpar{wang2025deepnotenotecentricdeepretrievalaugmented}   &45.0&41.5&20.5&52.0&52.0&55.0&44.3\\
\method{} &54.0 &61.5 &31.5 &68.8&62.0 &65.5&57.2\\
\bottomrule
\end{tabular}
}
\caption{Overall Performance of Different RAG Models. The \textbf{best} and \underline{second best} results are highlighted. In our experiments, we employ GLM-4.5 as a backbone mode.}
\label{tab:glm_overall}
\end{table*}
\begin{table*}[t]
\centering
\small
\resizebox{\linewidth}{!}{
\begin{tabular}{lccccccc}
\toprule
\textbf{Methods} & \textbf{HotpotQA} & \textbf{2WikiMQA} & \textbf{MuSiQue}& \textbf{Bamboogle} & \textbf{NQ} & \textbf{AmbigQA} & \textbf{Avg.} \\
\midrule
\rowcolor{gray!8}\multicolumn{8}{l}{\textbf{\textit{Inference Time Latency}}} \\
\midrule
DeepNote & 3.63 & 3.67 & 3.99 & 4.79  & 4.05 & 4.18 & 4.05
\\
\method{} (Parallel Filling) & 2.54&  1.69&     2.14&  1.72&2.60& 2.39 &2.18\\
\method{} & 3.82& 3.42& 3.88&  3.76& 4.38&  3.71&3.82\\
\midrule
\rowcolor{gray!8}\multicolumn{8}{l}{\textbf{\textit{Inference Performance}}} \\
\midrule
DeepNote &48.4 &47.2 &17.2 &39.2 &55.6 &55.9 &43.9 \\
\method{} (Parallel Filling) &45.9 &43.8 &18.8 &59.2 &56.4 &57.0 &46.9 \\
\method{} & 50.6 & 57.4 & 23.0 &\ 62.4 & 56.5 & 56.4 & 51.1\\
\bottomrule
\end{tabular}
}
\caption{Inference Time Latency. The unit of inference latency is seconds.}
\label{tab:infer_latency}
\end{table*}

\subsection{The Computation of Document Information Gain}\label{app:infogain}
The document information gain (DIG) metric serves to quantify the actual utility of external knowledge within RAG systems~\citep{DBLP:journals/corr/abs-2509-12765}.

Formally, for a question $q$, a knowledge representation $O$, and the ground truth $y$, DIG is defined as the difference in the LLM's generation confidence for the correct answer when the $K$ is included versus when it is excluded. Let $p_{\phi}(y \mid q,O)$ denote the conditional probability of the model generating the ground truth $y$ when knowledge representation $O$ is augmented into the context, and let $p_{\phi}(y \mid x)$ represent the probability of generating the ground truth $y$ without the augmentation of knowledge representation $O$. Then, the DIG is calculated as:
\begin{equation}
\label{eq:dig}
\small
\text{DIG}(O \mid q) = p_{\phi}(y \mid q, O) - p_{\phi}(y \mid q),
\end{equation}
where $\phi$ denotes the parameters of the LLM. Furthermore, in the calculation of DIG, previous work introduces two strategies specifically designed to mitigate length bias and ensure the capture of the strongest signals indicative of generation quality:

\paragraph{Sliding Window Smoothing.} To mitigate length bias in long sequences, a sliding window mechanism is utilized to smooth local probability fluctuations. For each token $t_i$ in the answer sequence $y$, its smoothed probability is calculated as:
\begin{equation}
\label{eq:smooth}
\small
p_{s}(t_i) = \frac{1}{W} \sum_{j=i-\lfloor W/2 \rfloor}^{i+\lfloor W/2 \rfloor} p(t_j),
\end{equation}
where $W$ denotes the window size and $p(t_j)$ represents the original token probability.

\paragraph{Token Importance Weighting.} To emphasize the core semantic information often encoded in initial tokens, a weighting scheme assigns higher weights to the first $k$ tokens. The final calibrated probability score is derived as:
\begin{equation}
\label{eq:smooth}
\small
p_{\phi}(y|x) = \prod_{i=1}^{k} (p_{s}(t_i))^{\omega_i \cdot \alpha} \cdot \prod_{j=k+1}^{|y|} (p_{s}(t_j))^{1-\alpha},
\end{equation}
where $\omega_i$ is the importance weight for the $i$-th token, and $\alpha$ is a hyperparameter controlling the emphasis on the initial segment.

\begin{figure}[t]
  \centering
    \subfigure[HotpotQA.]{
    \includegraphics[width=0.46\linewidth]{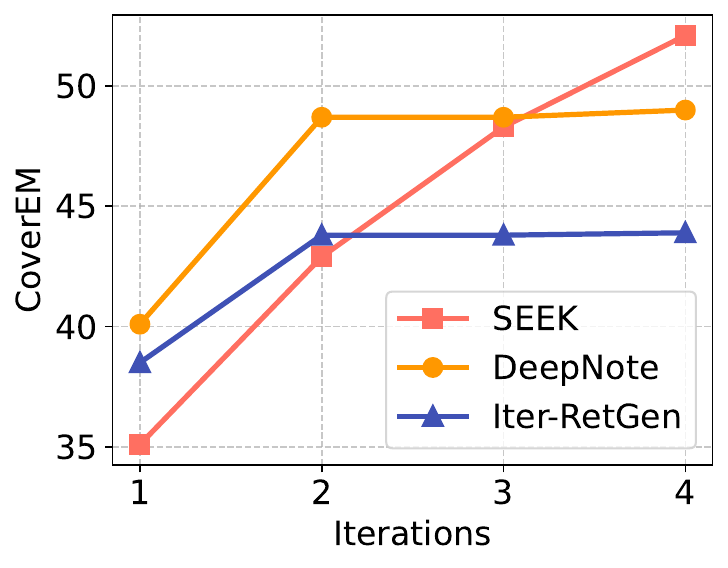}

  }
  \subfigure[MuSiQue.]{
    \includegraphics[width=0.46\linewidth]{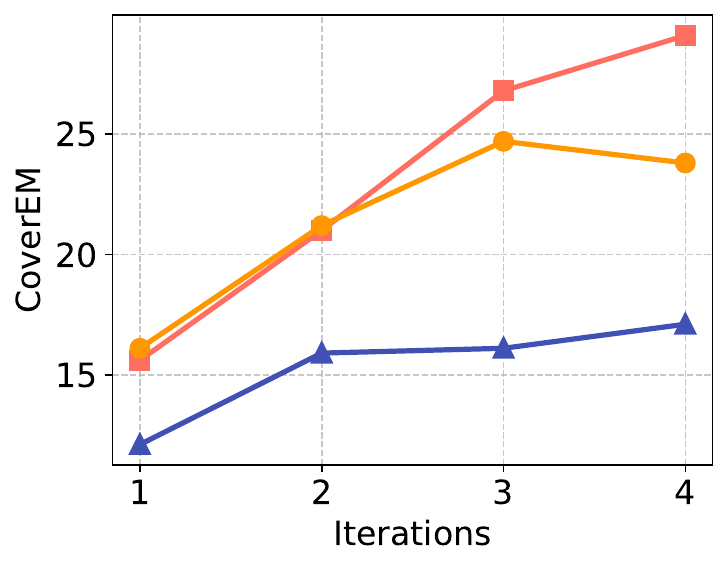}

  }
\caption{The Performance of Models Evolution across Iteration Rounds.}
  \label{fig:coverem_baohedu}
\end{figure}

\begin{figure}[t]
  \centering
  \subfigure[HotpotQA.]{
    \includegraphics[width=0.46\linewidth]{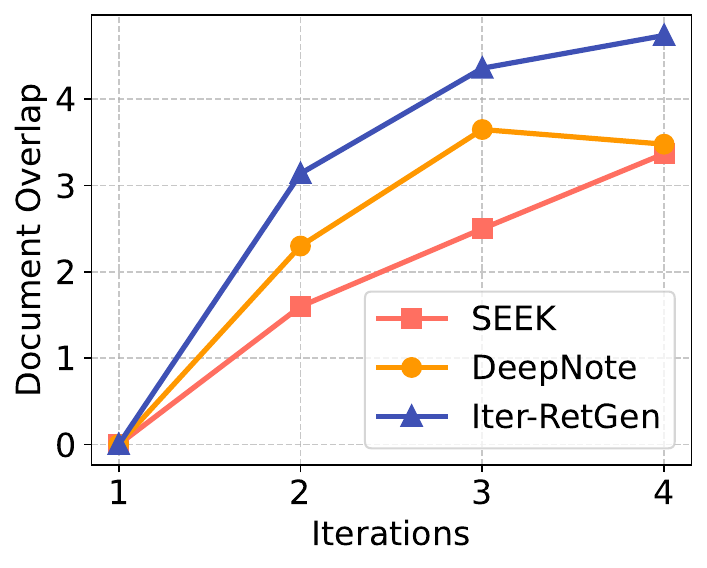}

  }
  \hfill
  \subfigure[MuSiQue.]{
    \includegraphics[width=0.46\linewidth]{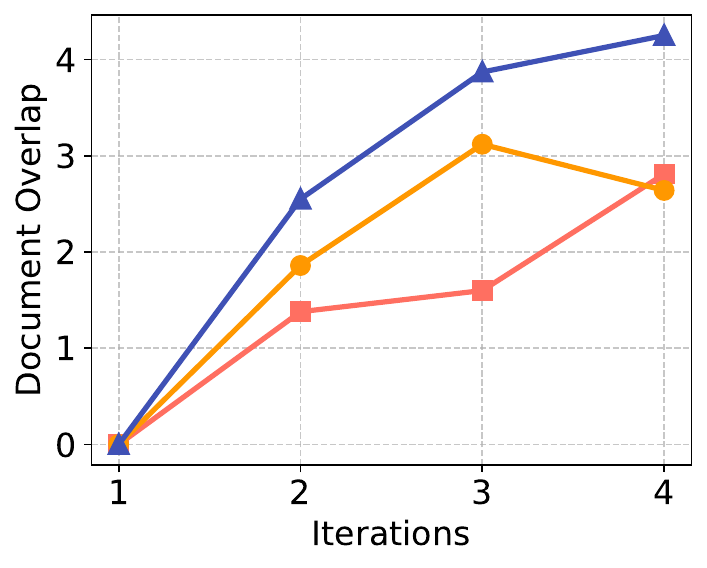}
  }
  \caption{The Document Overlap of Models Evolution across Iteration Rounds.}
  \label{fig:doc_baohedu}
\end{figure}
\subsection{Characteristics of Knowledge Construction across Different Methods}\label{sec:app:exp}

In this section, we further investigate the characteristics of different methods during the iterative knowledge accumulation process. We adopt Qwen3-32B as the backbone model for all experiments. We select questions that require four rounds of iterative retrieval for \method{}, DeepNote, and Iter-RetGen as the evaluation set.

As illustrated in Figure~\ref{fig:coverem_baohedu}, we first analyze how the performance of each method evolves throughout the iterative process. We observe that, as the number of iterations increases, the performance gains of Iter-RetGen and DeepNote rapidly plateau after the second round. This indicates that these methods encounter a bottleneck in continuously acquiring and integrating new knowledge across multiple iterations, making it difficult to further capture key evidence in subsequent rounds. In contrast, \method{} exhibits steady performance improvement as the iterative process progresses. This suggests that, under the guidance of the steering sketch, \method{} can progressively acquire more useful external knowledge and incorporate it into the completed sketch, thereby continuously improving model performance.

Furthermore, we analyze retrieval diversity in Figure~\ref{fig:doc_baohedu}, which shows the overlap between the documents retrieved in the current iteration and the cumulative set of documents retrieved in previous iterations. The results show that DeepNote and Iter-RetGen exhibit higher document overlap than \method{}, indicating that they are more likely to repeatedly retrieve similar evidence during iterative retrieval. In contrast, the lower document overlap of \method{} suggests that the steering sketch can guide the model to explore more diverse knowledge aspects across multiple iterations to better support answer generation.

\subsection{The Performance of \method{} on Closed-source LLM}\label{app:glm}
As shown in Table~\ref{tab:glm_overall}, we further evaluate \method{} with GLM-4.5 as the backbone model and compare it with representative baseline methods on knowledge-intensive tasks. We randomly sample 200 instances from each dataset for evaluation.

Overall, \method{} consistently outperforms all baselines under the closed-source LLM setting, achieving an average performance improvement of over 3\%. This demonstrates that the effectiveness of \method{} is not limited to open-source backbone models such as Qwen3-32B and Llama3.1-70B-Instruct. Instead, \method{} can also generalize to closed-source LLMs, suggesting that steering sketch-guided knowledge acquisition is a model-agnostic strategy for improving RAG performance.

\begin{table*}[t]
\centering
\small
\begin{tabular}{lccccccc}
\toprule
\textbf{\# Slots} & \textbf{HotpotQA} & \textbf{2WikiMQA} & \textbf{MuSiQue}& \textbf{Bamboogle} & \textbf{NQ} & \textbf{AmbigQA} & \textbf{Avg.} \\
\midrule
\rowcolor{gray!8}\multicolumn{8}{l}{\textbf{\textit{Qwen3-32B}}} \\
\midrule
$\leq$ 2 &0.4&1.0&0.4&1.6&0.1&0.0 &0.6 \\
= 3 &20.8&27.4&20.8&29.6&5.8&9.4 &18.9\\
= 4 &35.9&42.6&36.4&40.0&30.5&36.5 &37.0\\
= 5 &32.4&23.0&30.1&21.6&44.3&40.3 &31.9\\
= 6 &7.5&4.5&8.1&4.8&13.6&9.9&8.1 \\
$\geq$ 7 &3.0&1.5&4.2&2.4&5.7&3.9 &3.5 \\
\midrule
\rowcolor{gray!8}\multicolumn{8}{l}{\textbf{\textit{Llama3.1-70B-Instruct}}} \\
\midrule
$\leq$ 2 &2.5&4.0&4.1&3.2&1.0&1.1&2.7 \\
= 3 &44.9&53.0&45.1&52.8&38.6&47.9 &47.1\\
= 4 &40.6&37.3&37.2&34.4&36.1&35.8&36.9 \\
= 5 &10.4&5.6&11.4&9.6&19.4&13.0&11.6\\
= 6 &1.4&0.1&1.9&0.0&3.8&1.8& 1.4\\
$\geq$ 7 &0.2&0.0&0.3&0.0&1.1&0.4&0.3 \\
\bottomrule
\end{tabular}
\caption{Statistics of Steering Gist Counts in Steering Sketches.}
\label{tab:number_section}
\end{table*}

\subsection{Inference Time Latency}\label{app:latency}

In this section, we compare the inference latency of DeepNote, \method{}, and \method{} (Parallel Filling). \method{} (Parallel Filling) is an ablated variant introduced in Section~\ref{ablation}, which generates sub-queries for all unfilled slots in the initial steering sketch simultaneously and performs retrieval in parallel. We use Qwen3-32B as the backbone model.

As shown in Table~\ref{tab:infer_latency}, \method{} requires higher latency than \method{} (Parallel Filling), with an average latency of 3.82 seconds compared to 2.18 seconds. This additional cost mainly comes from the sequential sketch updating process, where \method{} iteratively generates sub-queries, retrieves documents, refines evidence, and fills the corresponding slots. Nevertheless, the latency does not increase linearly with the number of iterations: although \method{} typically performs around four iterations, its latency is only about 1.8$\times$ that of \method{} (Parallel Filling), rather than 4$\times$. This is because the parallel variant retrieves evidence for all slots at once and therefore needs to process a larger set of retrieved documents in a single context.

More importantly, the additional latency brings clear performance gains. Compared with \method{} (Parallel Filling), \method{} improves the average performance from 46.9 to 51.1, showing that sequential sketch updating enables the model to adjust subsequent sub-queries based on previously filled evidence. Compared with DeepNote, \method{} achieves better average performance while maintaining slightly lower average latency. These results suggest that \method{} provides a favorable trade-off between effectiveness and efficiency among iterative knowledge-construction RAG methods.


\subsection{Statistics of Steering Sketch Gist Count} \label{Statistics_of_page_slots}

In this section, we further investigate the number of steering gists initialized by \method{} and analyze their distribution across different datasets and backbone models. As shown in Table~\ref{tab:number_section}, the number of steering gists is mainly concentrated between 3 and 5. This indicates that, during steering sketch initialization, \method{} leverages the planning capability of the LLM to organize the question into a compact set of key knowledge aspects.

A closer comparison further shows that Qwen3-32B tends to initialize sketches with four or five steering gists, while Llama3.1-70B-Instruct more frequently produces sketches with three or four steering gists. This difference suggests that different backbone LLMs may exhibit different preferences in decomposing questions into knowledge aspects, leading to variations in the resulting steering sketch structure. Overall, the distribution demonstrates that \method{} constructs concise yet structured steering sketches, avoiding both overly coarse and overly fragmented knowledge decomposition.

\subsection{Case Studies of \method{}}

In this section, we present several case studies to further demonstrate the effectiveness of \method{}, as well as the iterative steering sketch construction process of \method{}. All cases are selected from the HotpotQA dataset.

\begin{figure*}[t]
\centering
\includegraphics[width=1\linewidth]{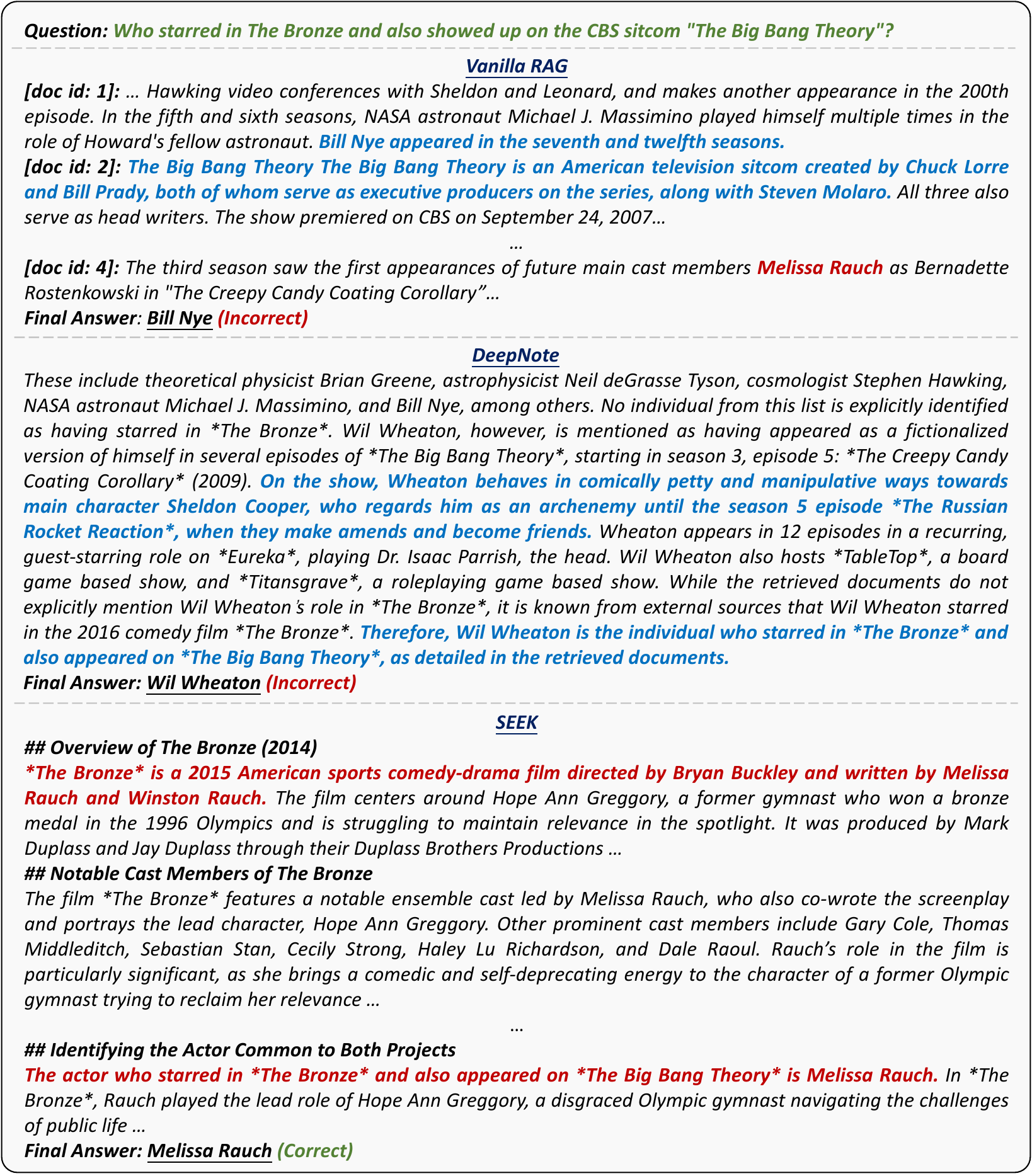}
\caption{Case Study of Different RAG Models. Text in \textbf{\textcolor[rgb]{0.7,0.1,0.1}{red}} represents document excerpts directly related to the knowledge representation and the question answering, while text in \textbf{\textcolor[rgb]{0.0,0.4,0.8}{blue}} represents the noisy content in the knowledge representation.}
\label{fig:case1}
\end{figure*}

As illustrated in Figure~\ref{fig:case1}, we compare the knowledge representations and final predictions generated by Vanilla RAG, DeepNote, and \method{}. Vanilla RAG and DeepNote incorrectly identify ``Bill Nye'' and ``Wil Wheaton'', respectively, as the correct actors, despite neither having starred in ``The Bronze''. In contrast, the steering sketch constructed by \method{} progressively organizes the retrieved evidence into semantically coherent steering gists. Specifically, \method{} successfully consolidates the key evidence that ``Melissa Rauch'' starred in ``The Bronze'' and also appeared in ``The Big Bang Theory''. This case demonstrates that the structured steering sketch enables \method{} to better organize and refine retrieved knowledge, thereby helping the LLM generate the correct answer.

\begin{figure*}[t]
\centering
\includegraphics[width=\linewidth]{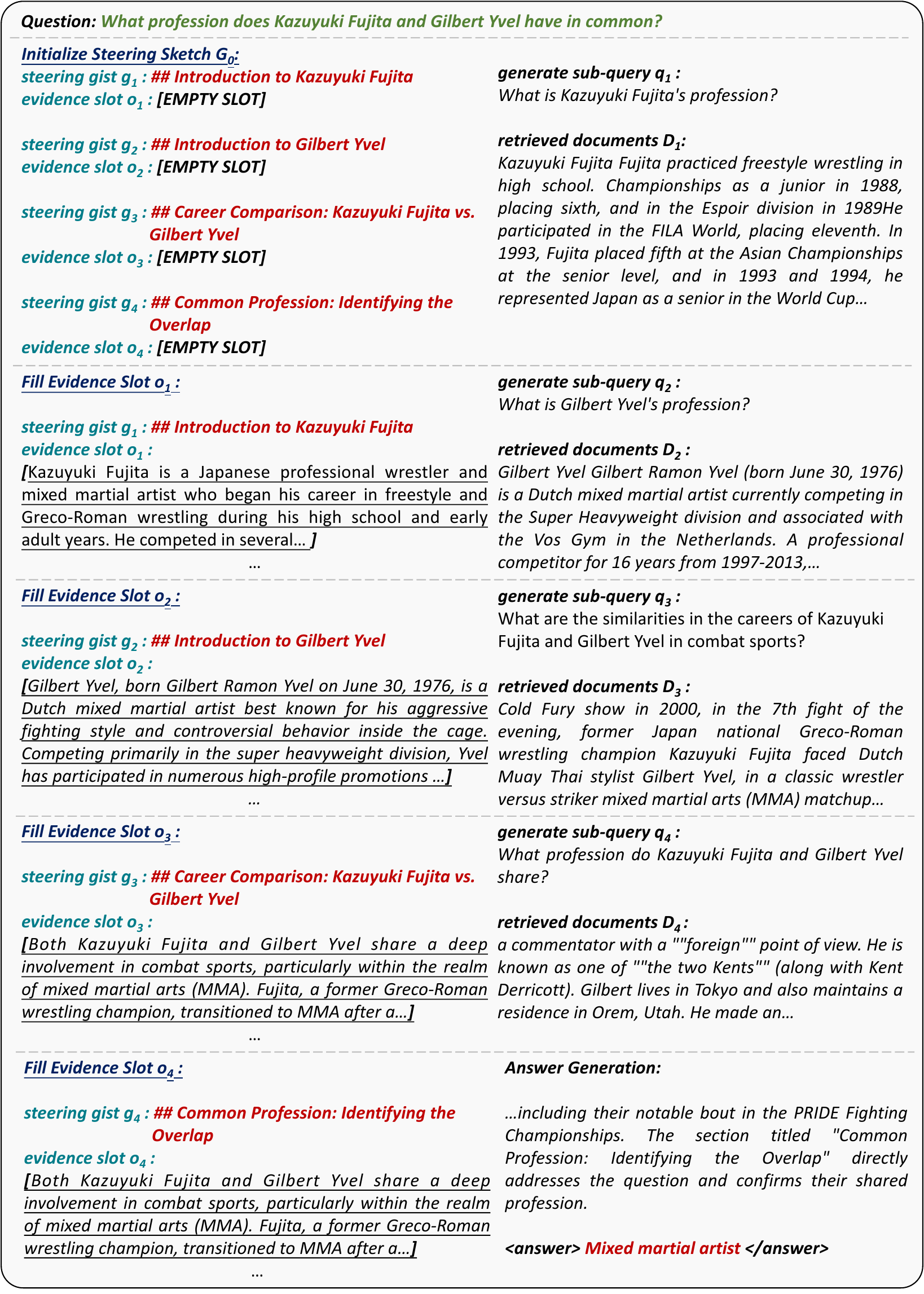}
\caption{Case Study of Steering Sketch Construction. Text in \underline{underlined} represents the knowledge evidence filled into the steering gist slots.}
\label{fig:case2}
\end{figure*}

Figure~\ref{fig:case2} further illustrates the iterative steering sketch construction process of \method{}. During the initialization stage, \method{} first constructs a structured steering sketch tailored to the query, where each steering gist corresponds to a distinct knowledge aspect associated with the question. These steering gists provide structured guidance for the subsequent iterative knowledge acquisition process. In Iterations 1 and 2, \method{} generates sub-queries to retrieve external documents and refines the retrieved knowledge into evidence for the corresponding steering gist slots. Subsequently, in Iterations 3 and 4, the generated sub-queries further explore the relationship between the two entities. The retrieved evidence reveals that the two subjects directly competed in a 2000 tournament. Accordingly, \method{} synthesizes this information within the ``Career Comparison'' steering gist, explicitly capturing that both subjects are deeply involved in the field of mixed martial arts (MMA). Finally, based on the completed steering sketch, \method{} correctly identifies the answer as ``Mixed martial artist'' and grounds its prediction using the accumulated evidence stored in the steering sketch. This case further demonstrates that the iterative steering sketch construction mechanism enables \method{} to progressively organize and integrate retrieved knowledge into a logically coherent representation for question answering.

\begin{figure*}[t]
\centering
\includegraphics[width=\linewidth]{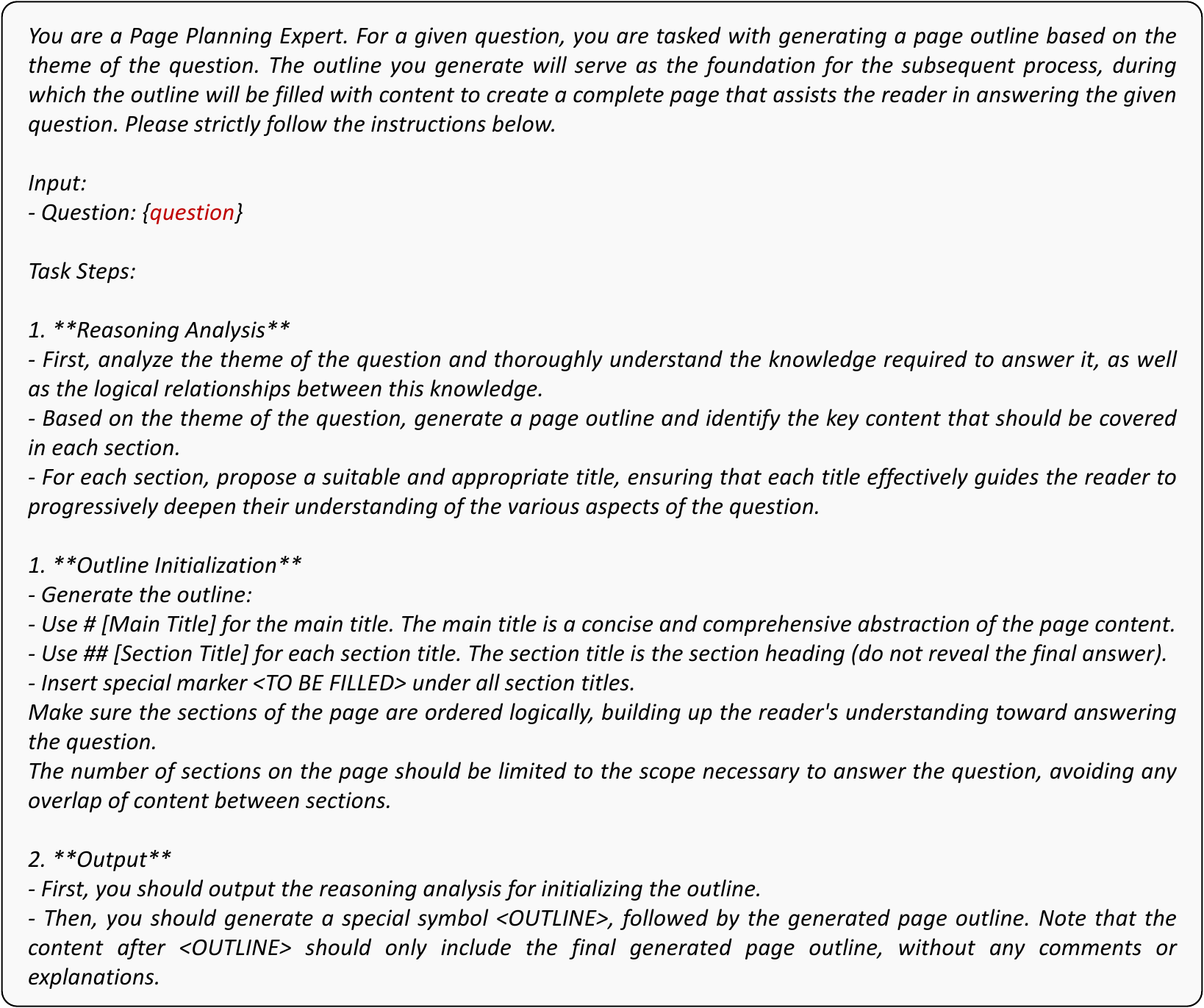}
\caption{Prompt Template for Initializing the Steering Sketch.}
\label{fig:prompt1}
\end{figure*}

\begin{figure*}[t]
\centering
\includegraphics[width=\linewidth]{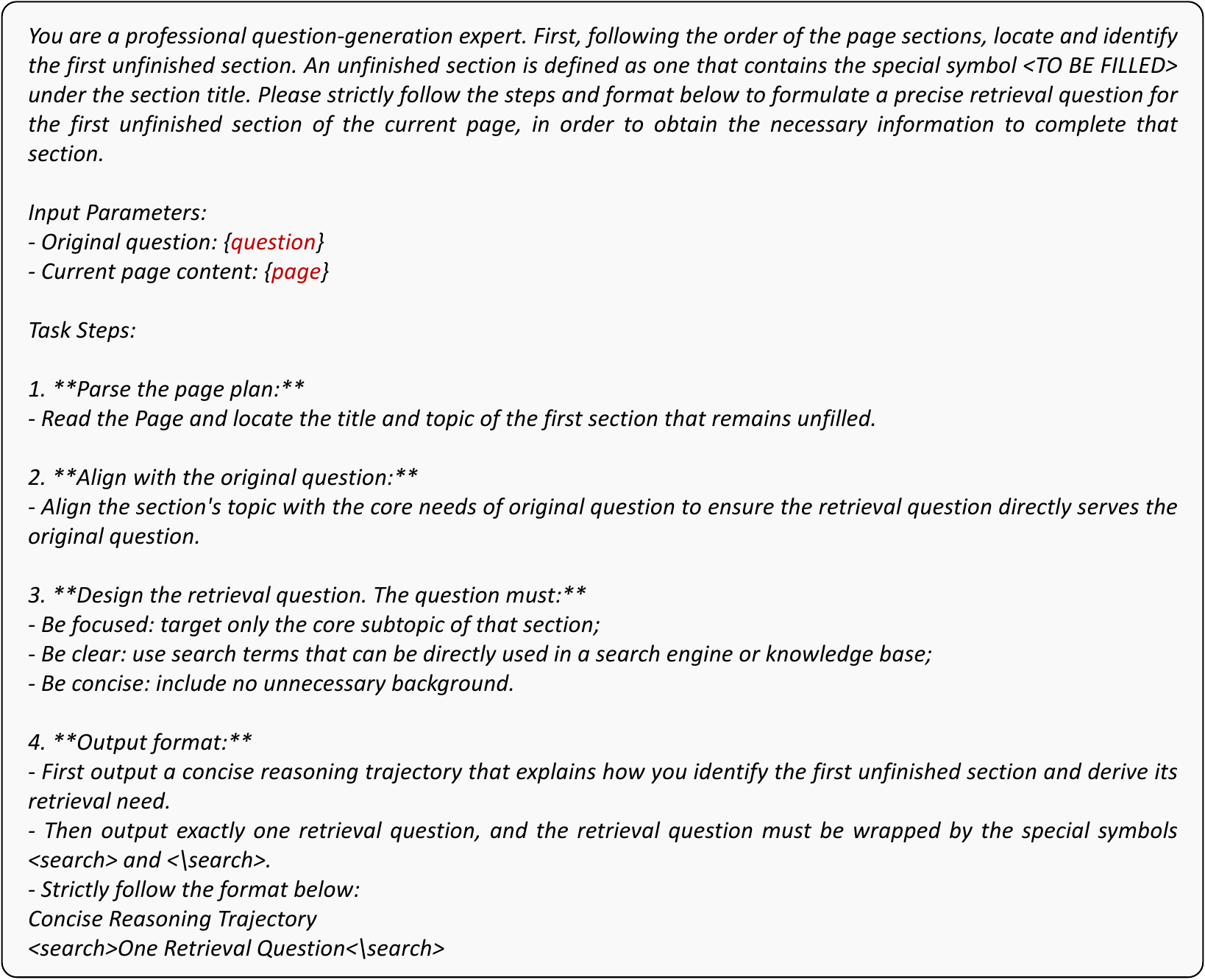}
\caption{Prompt Template for Generating Sub-queries.}
\label{fig:prompt2}
\end{figure*}
\begin{figure*}[t]
\centering
\includegraphics[width=\linewidth]{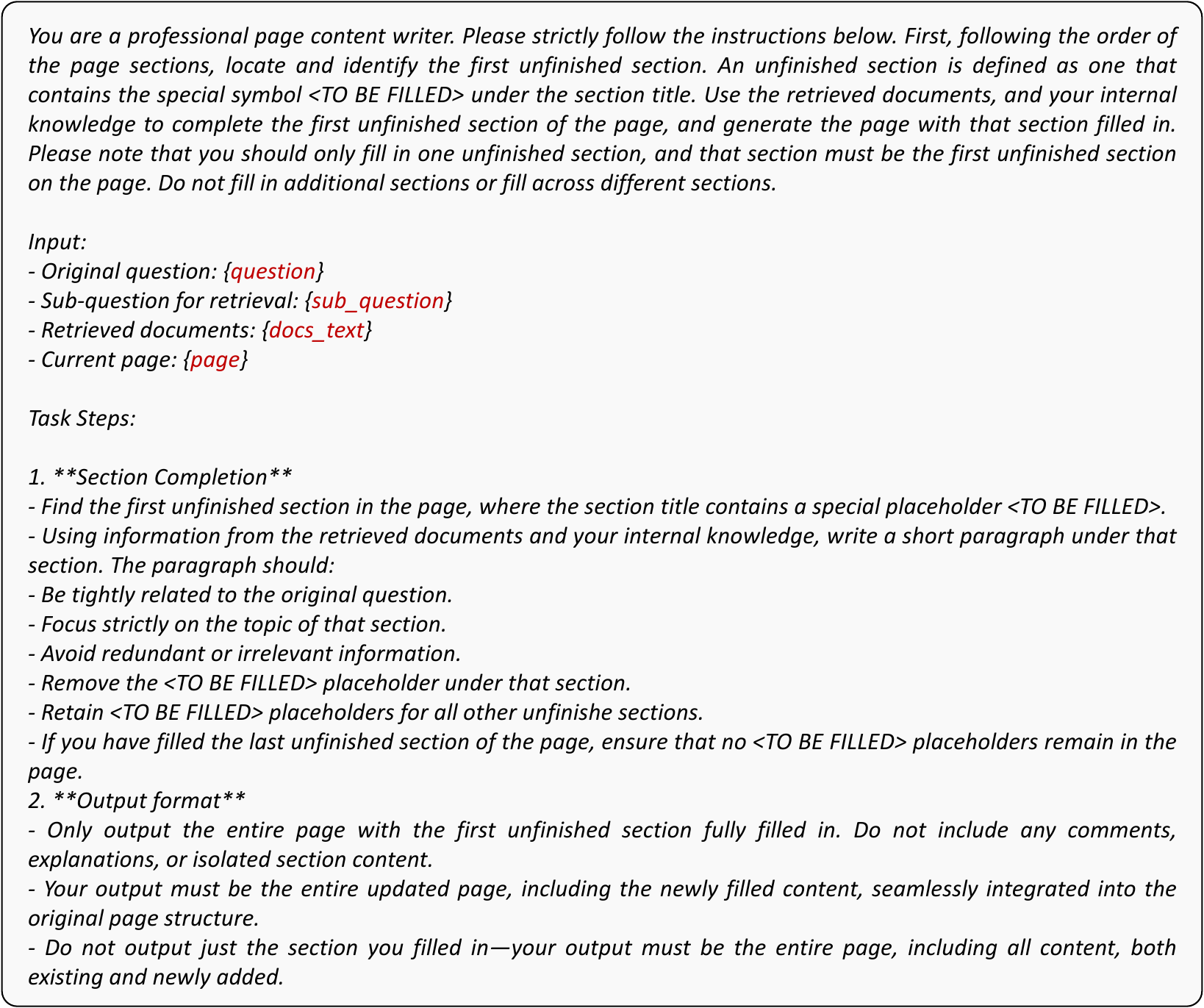}
\caption{Prompt Template for Filling Steering Gist Slots.}
\label{fig:prompt3}
\end{figure*}

\begin{figure*}[t]
\centering
\includegraphics[width=\linewidth]{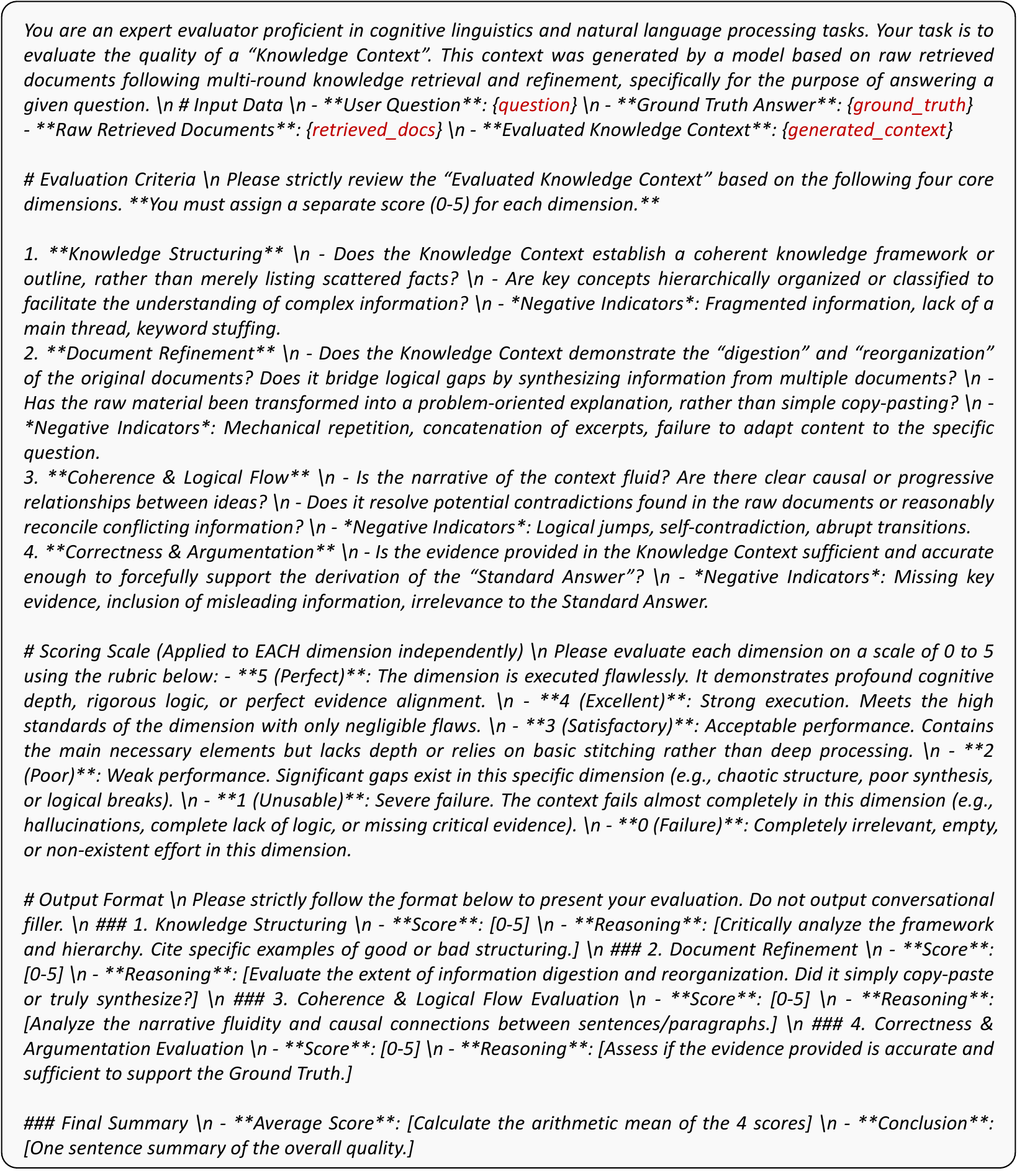}
\caption{Prompt Template for Evaluating Knowledge Representations.}
\label{fig:prompt4}
\end{figure*}

\subsection{Prompt Templates Used for \method{}} \label{app:prompt}

We provide the prompt templates used in the experiments of \method{}. 
Figure~\ref{fig:prompt1} presents the prompt for initializing the steering sketch from the input question. 
Figure~\ref{fig:prompt2} shows the prompt for generating a sub-query under the guidance of the current steering sketch. 
Figure~\ref{fig:prompt3} illustrates the prompt for refining retrieved documents into evidence and filling the corresponding steering gist slot in the steering sketch.
Figure~\ref{fig:prompt4} presents the prompt for evaluating the quality of knowledge representations generated by different models.

\end{document}